\documentclass[10pt,twocolumn,letterpaper]{article}

\usepackage{wacv}
\usepackage{times}
\usepackage{epsfig}
\usepackage{graphicx}
\usepackage{amsmath}
\usepackage{amssymb}
\usepackage{booktabs}

\usepackage{cite}
\usepackage{ctable}
\usepackage{hhline}
\usepackage{booktabs}
\usepackage{subcaption}
\usepackage{csquotes}
\usepackage{multirow}

\usepackage{color, colortbl}
\definecolor{Gray}{gray}{0.9}

\newcolumntype{L}[1]{>{\raggedright\let\newline\\\arraybackslash\hspace{0pt}}m{#1}}
\newcolumntype{C}[1]{>{\centering\let\newline\\\arraybackslash\hspace{0pt}}m{#1}}
\newcolumntype{R}[1]{>{\raggedleft\let\newline\\\arraybackslash\hspace{0pt}}m{#1}}

\def\wacvPaperID{161} % Enter the WACV Paper ID here

\wacvalgorithmstrack   % Uncomment this line if you are submitting to the Algorithms Track.
\wacvfinalcopy % *** Uncomment this line for the final submission

\ifwacvfinal
\usepackage[breaklinks=true,bookmarks=false]{hyperref}
\else
\usepackage[pagebackref=true,breaklinks=true,colorlinks,bookmarks=false]{hyperref}
\fi

\begin{document}

%%%%%%%%% TITLE
\title{Learnable Human Mesh Triangulation for 3D Human Pose and Shape Estimation}

\iffalse
\author{Sungho Chun\\
ECE, Kwangwoon University, Korea\\
{\tt\small firstauthor@i1.org}
\and
Sungbum Park\\
NCSOFT, Korea\\
First line of institution2 address\\
{\tt\small secondauthor@i2.org}
\and
Ju Yong Chang\\
ECE, Kwangwoon University, Korea\\
{\tt\small secondauthor@i2.org}
}
\fi

\iftrue
\author{
Sungho Chun$^1$ \hspace{1.0cm} Sungbum Park$^2$ \hspace{1.0cm} Ju Yong Chang$^1$\\
$^1$Dept of ECE, Kwangwoon University, Korea\hspace{0.5cm}
$^2$NCSOFT, Korea\\ 
{\small \texttt {\{asw9161,jychang\}@kw.ac.kr},\ \ \texttt {spark0916@ncsoft.com}}
}
\fi

\maketitle
\thispagestyle{empty}

%%%%%%%%% ABSTRACT
\begin{abstract}
Compared to joint position, the accuracy of joint rotation and shape estimation has received relatively little attention in the skinned multi-person linear model (SMPL)-based human mesh reconstruction from multi-view images. The work in this field is broadly classified into two categories. The first approach performs joint estimation and then produces SMPL parameters by fitting SMPL to resultant joints. The second approach regresses SMPL parameters directly from the input images through a convolutional neural network (CNN)-based model. However, these approaches suffer from the lack of information for resolving the ambiguity of joint rotation and shape reconstruction and the difficulty of network learning. To solve the aforementioned problems, we propose a two-stage method. The proposed method first estimates the coordinates of mesh vertices through a CNN-based model from input images, and acquires SMPL parameters by fitting the SMPL model to the estimated vertices. Estimated mesh vertices provide sufficient information for determining joint rotation and shape, and are easier to learn than SMPL parameters. According to experiments using Human3.6M and MPI-INF-3DHP datasets, the proposed method significantly outperforms the previous works in terms of joint rotation and shape estimation, and achieves competitive performance in terms of joint location estimation.
\end{abstract}
% A two-stage method was proposed to solve the aforementioned problems.

%%%%%%%%% BODY TEXT
%-------------------------------------------------------------------------
\section{Introduction}
\label{sec:intro}

\begin{figure}[t]
\centering
\includegraphics[width=\linewidth]{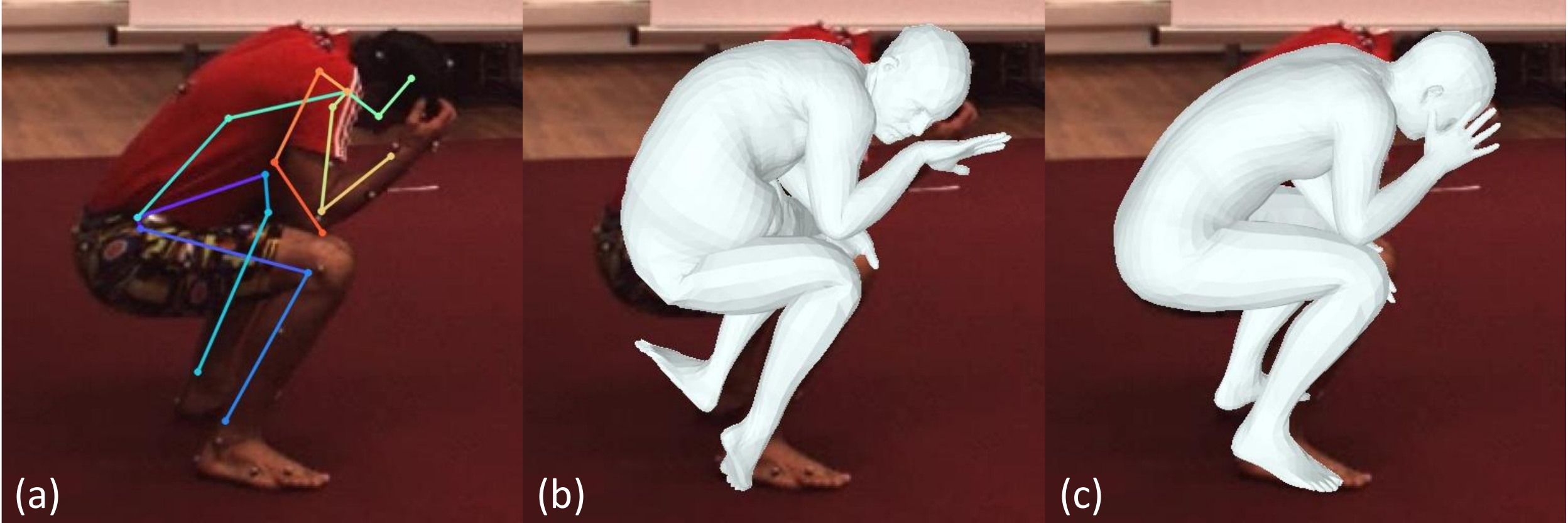}
\vspace*{-4mm}
\caption{Results for (a) joint position estimation, (b) joint fitting, and (c) surface fitting are visualized. Joint fitting and surface fitting indicate that SMPL is fitted to the estimated joint set and vertex set, respectively.}
\label{fig1}
\vspace*{-4mm}
\end{figure}

Human pose estimation from single or multi-view images is a long-standing computer vision problem. In many studies~\cite{2019_LT,2020_He_ET,2019_qiu,2018_Sun}, the human pose is simply represented as a set of 3D coordinates of the body joints. Compared to joint coordinate, human joint rotation and shape estimation has not received much attention. However, when 3D joint coordinates as well as joint rotations and human shape information are available together, the body of a person can be better described, as shown in Fig.~\ref{fig1}(a) and (c). The estimated joint and shape information can also be used for human part segmentation~\cite{2018_Kanazawa} and detailed human mesh reconstruction~\cite{zhu2019detailed, zhu2021detailed}.

The skinned multi-person linear model (SMPL)~\cite{2015_SMPL} is frequently used for multi-view human mesh reconstruction methods~\cite{2020_Shin,lightcap2021,wang2021mvp,airpose_2022arxiv}, which can acquire joint rotations and human shape as well as joint coordinates. Among the methods, the most similar to our proposed method is~\cite{lightcap2021}. This method first estimates 3D joints from multi-view images and then additionally computes joint rotation and shape information by fitting the SMPL-X~\cite{SMPLify_X_2019_Pavlakos} model to the 3D joints. However, this fitting framework heavily relies on regularization because joint coordinates do not provide enough information to resolve the ambiguity in the estimation of joint rotation and shape information. Nevertheless, the lack of such information can degrade joint rotation and human shape estimation performance, as shown in Fig.~\ref{fig1}(b). The convolutional neural network (CNN)-based model proposed in~\cite{2020_Shin} directly regresses SMPL pose and shape parameters from input multi-view images. However, the mapping function from the input image to the SMPL parameter is highly non-linear~\cite{2020_I2L}, which makes learning the model difficult. 
%(3) Also, in the case of the multi-view human mesh reconstruction task, it is difficult to learn a robust model in various environments because there are very few datasets including the ground-truth mesh. Therefore, it is important to design an algorithm that performs well even on datasets not used for training. However, in existing works, their generalization performance was not sufficiently verified because the same dataset was generally used for training and testing.

In this paper, we propose a \emph{Learnable human Mesh Triangulation (LMT)} method for SMPL-based human mesh reconstruction from sparse multi-view images. The proposed method can solve the above two problems. LMT first estimates human surface vertex coordinates, not human joints, from the input multi-view images, and then fits the SMPL model to the resultant vertices. Such surface vertex coordinates provide strong constraints on joint rotation and human shape, which can help resolve the ambiguity problem. Also, many previous works~\cite{fabbri2020compressed, bulat2017far, sun2019deep, WangSCJDZLMTWLX19, wei2016cpm, Luo_2021_CVPR} verified that heatmap-based keypoint estimation can be easily learned through CNNs, especially fully convolutional networks. Our basic idea is to extend this heatmap-based keypoint estimation framework to SMPL mesh vertex estimation, which can solve the non-linearity problem in direct SMPL parameter regression.
% Our basic idea is to extend this heatmap-based keypoint estimation framework to SMPL mesh vertex estimation, which can solve the non-linearity problem in direct SMPL parameter regression.

%Also, in the proposed LMT method, the geometry information obtained from the single-view mesh reconstruction method is used to predict the vertex coordinates. According to our cross-dataset evaluation, this helps to improve generalization performance by alleviating the multi-view inconsistency problem and (3) improving robustness.

%To reconstruct SMPL-based human mesh vertices, we adopt \emph{Learnable Triangulation of human pose (LT)}~\cite{2019_LT}, a heatmap-based human joint estimation model.
To reconstruct SMPL-based human mesh vertices, we extend \emph{Learnable Triangulation of human pose (LT)}~\cite{2019_LT}, the heatmap-based method for estimating sparse joints to dense vertices. However, the application of LT to mesh vertices is non-trivial and raises two issues to be overcome. The first is high computational complexity. LT generates 3D heatmaps to estimate body joints. No problem is observed in the case of sparse joints (e.g., $\sim$20 for Human3.6M~\cite{2014_H36M} and MPI-3DHP-INF~\cite{2017_3DHP}). In contrast, the use of a 3D heatmap may cause excessive GPU memory usage in the case of dense mesh vertices (e.g., 6890 for SMPL). However, the optimization process used to obtain SMPL parameters in the proposed method does not require full-vertices. Rather, estimating appropriately sampled sub-vertices can improve the performance of the model while solving the computational issue, which is proven through our experiments.

% To reconstruct SMPL-based human mesh vertices, we adopt \emph{Learnable Triangulation of human pose (LT)}~\cite{2019_LT}, a heatmap-based human joint estimation model.

The second issue is the inconsistency between multi-view features. In our method, multi-view features are aggregated in each voxel after being unprojected into 3D space. In the case of voxel on the human surface, multi-view features aggregated into the voxel must be consistent. However, occlusion can lead to inconsistency between aggregated multi-view features, which makes vertex coordinate estimation difficult. To alleviate this problem, we propose to utilize the visibility information obtained from the single-view mesh reconstruction method. The basic idea is to use visibility information to increase the dependence of a certain voxel on features obtained from visible views and reduce the dependence on features obtained from invisible views. We experimentally show that utilizing visibility information alleviates the multi-view inconsistency problem and improves mesh reconstruction performance.

The contributions of this paper can be summarized as follows:
\begin{itemize}
\item We quantitatively and qualitatively prove that fitting the SMPL model to human surface vertices rather than human body joints leads to better mesh reconstruction results in terms of joint rotation and human shape.
\item We show that the computational issue that makes it difficult to extend the heatmap-based framework to SMPL mesh vertices can be resolved through sub-vertices estimation, which also brings additional performance gain.
\item Per-vertex visibility information is utilized to consider the consistency of multi-view features. Moreover, cross-dataset experiments show that the use of visibility improves the generalization performance of our model.
\item Extensive experiments using Human3.6M and MPI-INF-3DHP datasets prove that the ideas of sub-vertices estimation and per-vertex visibility are effective. Consequently, the proposed framework outperforms previous methods in terms of joint rotation and human shape while showing competitive results in terms of 3D joint coordinates.
\end{itemize}

%-------------------------------------------------------------------------
\section{Related Work}
\label{sec:related_work}

\subsection{Multi-view Joint Estimation}

Many methods~\cite{2019_LT, 2019_qiu, 2020_He_ET, 2018_Tome, pavlakos17harvesting, 2018_Kadkhodamohammadi, Wu_2021_ICCV} have been proposed to estimate the 3D human pose in the form of joint coordinates from the input multi-view images. Among the methods for estimating the pose of a single person, the one most similar to our work is LT~\cite{2019_LT}. LT aggregates 2D features extracted from multi-view images in 3D voxel space and then applies the 3D convolution to the aggregated feature to estimate 3D pose. However, the final LT output is the 3D joint locations without joint rotation information. In contrast, in our method, the SMPL parameters are estimated, which enables a richer reconstruction of the human body, including joint rotations and human shape.

\begin{figure*}[t]
\centering
\includegraphics[width=\linewidth]{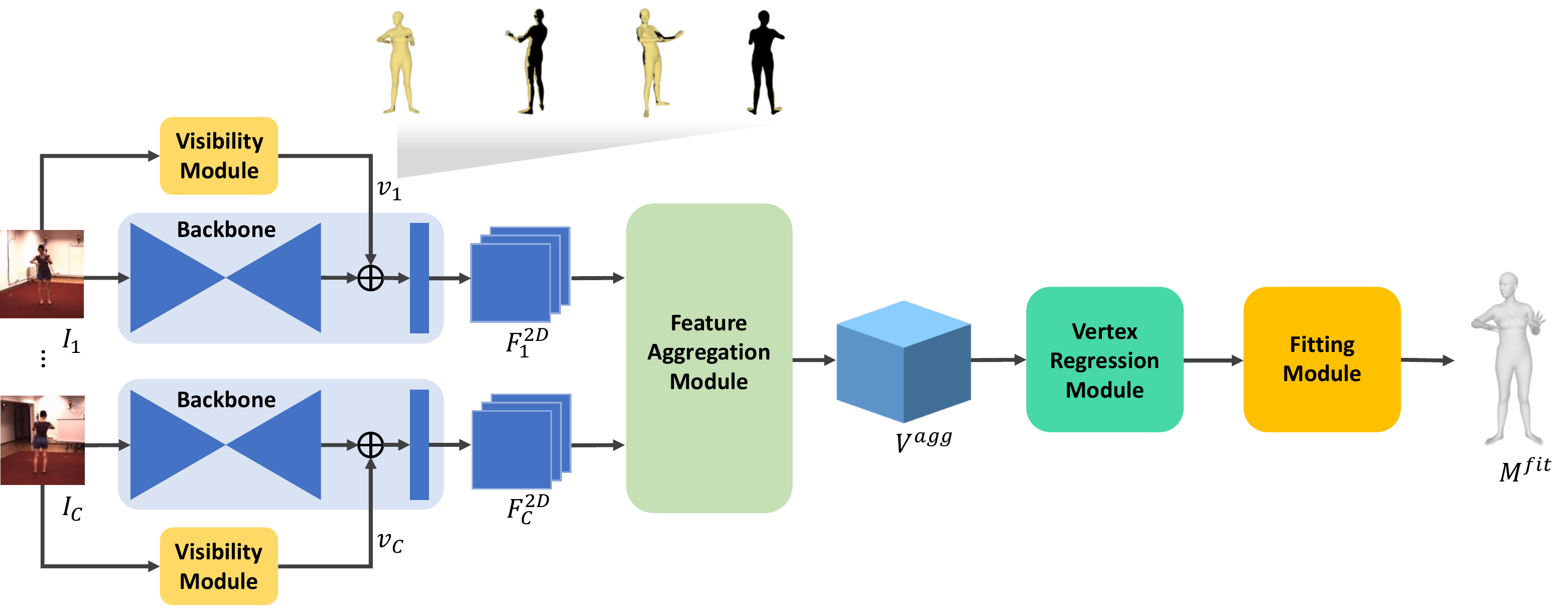}
\vspace*{-4mm}
\caption{\textbf{Overall pipeline of the proposed method.} Visible vertices in the visibility map are colored in gold. $\oplus$ denotes concatenate operation.}
%The input images $I_1,...,I_C$ are computed as a per-vertex visibility map $v_1,...,v_C$ through the visibility module. Visible vertices in the visibility maps are colored in gold. Visibility augmented image features $F^{2D}_{1},...,F^{2D}_{C}$ are calculated from the images and visibility maps input to the backbone. $\{F^{2D}_{c}\}_{c=1}^{C}$ input to the feature aggregation module are unprojected into a global 3d cuboid, and the generated $C$ unprojected features $V^{unproj}_{c}$ are aggregated into $V^{agg}$. Vertex regression module estimates subsampled human vertices $M$ from $V^{agg}$. The fitting module outputs the SMPL parameter by fitting the SMPL model to $M$.
%  $V^{agg}$ is calculated from $\{F^{2D}_{c}\}_{c=1}^{C}$ input to the feature aggregation module.
\label{fig2}
\vspace*{-4mm}
\end{figure*}

\subsection{Multi-view Joint and Shape Estimation}

Many studies~\cite{2020_Shin, lightcap2021, leroy2020smply, airpose_2022arxiv, wang2021mvp, zheng2021deepmulticap} have been conducted to estimate joint rotations or human shape as well as joint coordinates from input multi-view images. For SMPL and SMPL-X parameter estimation, the model is fitted to the predicted 3D joints in~\cite{lightcap2021}, and the 3D joints are fed into the feedforward network in~\cite{wang2021mvp}. In contrast, our method estimates SMPL parameters using 3D mesh vertices rather than 3D joints. Since the human surface provides richer information than joint coordinates for joint rotation and human shape estimation, our method can reconstruct rotation and shape more accurately than joint-based methods~\cite{lightcap2021, wang2021mvp}. In~\cite{leroy2020smply}, the Mannequin dataset~\cite{mannequin_2019cvpr} is used to train a model that robustly predicts SMPL parameters in an in-the-wild environment. The dataset provides videos of static humans captured by a dynamic camera. The method in~\cite{leroy2020smply} performs 3D joint estimation by applying the structure-from-motion (SfM) algorithm to the input video. However, the SfM method is generally difficult to apply to sparse multi-view environments, e.g., Human3.6M and MPI-INF-3DHP datasets, which are the focus of this work. The geometry of a clothed human is reconstructed in~\cite{zheng2021deepmulticap} and multiple images obtained from a dynamic camera are used in~\cite{airpose_2022arxiv}. Their goals and settings are different from our work.

In~\cite{2020_Shin}, an existing work with the same goal as ours, SMPL parameters are directly regressed from multi-view images through a CNN model. However, learning the network in this method is difficult due to the high non-linearity of the regression function~\cite{2018_Kanazawa, 2020_I2L, 2019_Kolotouros_ICCV, 2019_Kolotouros_CVPR, 2020_Choi, 2021_Lin_CVPR, 2021_Lin_ICCV}. Therefore, our method learns a keypoint estimation network based on heatmap regression rather than parameter regression, and then obtains SMPL parameters by fitting SMPL to human mesh vertices predicted by the network.

%-------------------------------------------------------------------------
\section{Proposed Method}
\label{sec:proposed_method}

\subsection{Overview of the Proposed Method}

We propose a method (i.e., LMT) to estimate the SMPL-based 3D mesh of a single person from multi-view images obtained by $C$ calibrated cameras. Fig.~\ref{fig2} shows the overall pipeline of the proposed method, which consists of the visibility module, CNN backbone, feature aggregation module, vertex regression module, and fitting module. The visibility module estimates per-vertex visibility $v_{c}\in{}\mathbb{R}^{N}$ for subsampled mesh from every single image $I_{c}\in{\mathbb{R}^{H_{0}\times{}W_{0}\times{}3}}$, where $N$ denotes the number of subsampled vertices. The CNN backbone computes visibility augmented image features $F_{c}^{2D}\in{\mathbb{R}^{H\times{}W\times{}K}}$ from the input multi-view image $I_c$ and per-vertex visibility $v_c$. The feature aggregation module unprojects the input image feature $F_{c}^{2D}$ into the 3D global voxel space to generate $C$ volumetric unprojected features $V_{c}^{unproj}\in{\mathbb{R}^{64\times{}64\times{}64\times{}K}}$, then aggregates the unprojected features $\{V_{c}^{unproj}\}_{c=1}^{C}$ to produce the volumetric aggregated feature $V^{agg}\in{\mathbb{R}^{64\times{}64\times{}64\times{}K}}$. The vertex regression module generates 3D vertex coordinates $M\in{\mathbb{R}^{N\times{}3}}$ of sub-sampled mesh from the aggregated feature $V^{agg}$ using 3D convolution and soft-argmax operation~\cite{2018_Sun}. The fitting module outputs the final joint coordinates, rotations, and shape information by fitting the SMPL model to the 3D vertex coordinates $M$ from the vertex regression module.

\subsection{Visibility Module}

The visibility module calculates the per-vertex visibility map $v_c$ from the single-view image $I_c$. We implement the visibility module using the I2L-MeshNet~\cite{2020_I2L}, one of the state-of-the-art single-view human mesh reconstruction methods, and the general visibility computation algorithm~\footnote{https://github.com/MPI-IS/mesh}. The detailed procedure is as follows. We first feed a single image $I_c$ into the I2L-MeshNet and obtain the human mesh defined in the human-centered coordinate system of which the origin is defined as the pelvis joint. However, the visibility computation algorithm requires camera coordinates of the human mesh. Therefore, the algebraic triangulation method~\cite{2019_LT} is used to estimate the pelvis joint. The camera coordinates of the estimated pelvis joint are used to transform the human mesh obtained by I2L-MeshNet into the camera coordinate system. The visibility computation algorithm is then used to obtain the visibility map for full-vertices ${v_c^{full}}\in{\mathbb{R}^{6890}}$. To prevent overfitting of the proposed model, we apply additional mesh subsampling~\cite{COMA_ECCV18} to $v_c^{full}$ and use the resultant per-vertex visibility map $v_c$ of sub-vertices for subsequent processes.
% We first feed a single image $I_c$ into I2L-MeshNet and obtain the human mesh defined in the human-centered coordinate system of which origin is defined as the pelvis joint.
% Therefore, we use the algebraic triangulation method~\cite{2019_LT} to estimate the pelvis joint.

\subsection{Backbone}

The CNN backbone outputs the visibility augmented image features $\{F_c^{2D}\}_{c=1}^{C}$ from input multi-view images $\{I_c\}_{c=1}^{C}$ and per-vertex visibility $\{v_c\}_{c=1}^{C}$. To construct the proposed backbone, according to~\cite{2019_LT}, we remove the last classification and pooling layers of ResNet-152~\cite{2016_He} pretrained on COCO~\cite{coco_eccv2014_microsoft} and MPII~\cite{MPII_Andriluka_2014_CVPR}, and then add three deconvolution layers and a $1\times{}1$ convolution layer to the back of the network. The last deconvolution layer of the backbone creates an intermediate feature $F_c^{deconv}\in{}\mathbb{R}^{H\times{}W\times{}256}$. After $v_c$ is extended to the spatial axis, it is concatenated with the intermediate feature $F_c^{deconv}$. An additional $1\times{}1$ convolution is applied to the concatenated feature to generate the visibility augmented image feature $F_c^{2D}$.

\subsection{Feature Aggregation Module}

In the feature aggregation module, the 2D feature $F_c^{2D}$ from the backbone is unprojected into a cuboid defined in 3D world space to create a volumetric unprojected feature $V_c^{unproj}$. The volumetric aggregated feature $V^{agg}$ is then calculated through the aggregation of the unprojected features $\{V_c^{unproj}\}_{c=1}^{C}$. In the proposed method, the estimation of the vertex coordinates of the subsampled mesh $M$ depends on the unprojected 3D features in the cuboid. Therefore, the location and size of the cuboid should be set so that the cuboid contains the target human subject. Consequently, a cuboid with a side length of 2.0 m, centering on the pelvis of the target subject, is created.

The construction process of the unprojected feature $V_c^{unproj}$ through the unprojection of $F_c^{2D}$ is as follows. We first project the 3D coordinates of the cuboid voxels $V^{coords}\in{}\mathbb{R}^{64\times{}64\times{}64\times{}3}$ into the 2D image plane of each view using the camera projection matrix and obtain the 2D image coordinates $V_c^{proj}\in{}\mathbb{R}^{64\times{}64\times{}64\times{}2}$. Next, bilinear sampling is used to extract 2D features corresponding to each location of $V_c^{proj}$ from $F_c^{2D}$, and, consequently, $V_c^{unproj}$ is obtained:
\begin{equation}
\label{eq:unprojection}
    V_c^{unproj}=F_c^{2D}\{V_c^{proj}\},
\end{equation}
where $\{\cdot\}$ denotes bilinear sampling. Then $C$ unprojected features in 3D world space are aggregated using 3D softmax operation~\cite{2019_LT}. This can be written as:
\begin{equation}
\label{eq:aggregate_1}
    V^{agg}=\sum_{c=1}^{C}(d_{c}\odot{}V_{c}^{unproj}),
\end{equation}
\begin{equation}
\label{eq:aggregate_2}
    d_{c}=\frac{\exp(V_c^{unproj})}{\sum_{c=1}^{C}\exp(V_c^{unproj})},
\end{equation}
where $d_c\in{\mathbb{R}^{64\times{}64\times{}64\times{}K}}$ and $\odot$ denote the confidence weight and element-wise multiplication, respectively.

\subsection{Vertex Regression Module}

The vertex regression module with encoder-decoder structure composed of 3D convolution generates the vertex coordinates of the subsampled mesh $M$ from the input aggregated feature $V^{agg}$. The encoder first computes a 3D feature with $2\times{}2\times{}2$ resolution and 128 channel dimension from $V^{agg}$, which is fed into the decoder to output a volumetric feature $V\in{}\mathbb{R}^{64\times{}64\times{}64\times{}32}$. Next, a $1\times{}1\times{}1$ 3D convolution is applied to $V$ to produce 3D heatmaps $H^{3D}\in{}\mathbb{R}^{64\times{}64\times{}64\times{}N}$ for the subsampled vertices. Details of the proposed encoder-decoder are presented in the Supplementary material.

A 3D soft-argmax operation is used to obtain vertex coordinates $M$ from the 3D heatmaps $H^{3D}$:
\begin{equation}
\label{eq:softargmax_1}
    \tilde{H}_{n}^{3D}=\frac{\exp(H_{n}^{3D})}{\sum_{i,j,k}\exp(H_{n}^{3D}(i,j,k))},
\end{equation}
\begin{equation}
\label{eq:softargmax_2}
    M_{n}=\sum_{i,j,k}r\cdot{}\tilde{H}_{n}^{3D}(i,j,k),
\end{equation}
where $r=[r_{i},r_{j},r_{k}]$ denotes the world coordinate vector of the voxel with indices $(i,j,k)$ in the 3D heatmap. $H_{n}^{3D}$, $\tilde{H}^{3D}$, and $M_{n}$ denote the $n$-th channel of the 3D heatmap, the normalized 3D heatmap, and the $n$-th row vector of $M$, respectively.
% That is, $r_{x}\in{}\{1,\ldots,W\}$, $r_{y}\in{}\{1,\ldots,H\}$, and $r_{z}\in{}\{1,\ldots,D\}$ represent the x, y, and z-coordinates of the 3D heatmap, respectively.

To train the proposed network, an L1 loss is applied to the vertices generated by the vertex regression module:
\begin{equation}
\label{eq:loss}
    \mathcal{L}_{M}=\frac{1}{N}\sum_{n=1}^{N}\|M_{n}-M_{n}^{*}\|_{1},
\end{equation}
where $M^{*}$ denotes the ground-truth mesh.

\subsection{Fitting Module}

The fitting module is used to acquire the SMPL parameters corresponding to the vertex coordinates $M$ generated by the vertex regression module. Fitting module is based on optimization according to the existing works~\cite{mosh_Loper_2014, moshpp_AMASS_2019, MOJO_CVPR_2021, lightcap2021, SMPLify_X_2019_Pavlakos} and optimization parameters $\Theta=\{z\in{}\mathbb{R}^{32},R\in{}\mathbb{R}^{6},\beta\in{}\mathbb{R}^{10},t\in{}\mathbb{R}^3\}$ contains VPoser’s latent code $z$, global rotation with continuous representation~\cite{Zhou_2019_CVPR} $R$, shape parameter $\beta$, and global translation $t$. From the latent code, VPoser $\mathcal{V}(\cdot)$ calculates the SMPL pose parameter $\theta=\mathcal{V}(z)\in{}\mathbb{R}^{69}$, which is fed into the SMPL decoder $\mathcal{M}(\cdot)$ together with $R$, $\beta$, and $t$ to produce the SMPL mesh $M^{fit}=\mathcal{M}(\theta,R,\beta,t)\in{}\mathbb{R}^{6890\times{}3}$. The SMPL mesh is transformed into sub-vertices $M_{sub}^{fit}=sub(M^{fit})\in{}\mathbb{R}^{N\times{}3}$ by the mesh coarsening function~\cite{COMA_ECCV18} $sub(\cdot)$. The fitting module updates $\Theta$ iteratively to reduce the difference between the sub-vertices of the fitted mesh $M_{sub}^{fit}$ and the regressed vertices $M$.

The cost function for fitting is defined as follows:
\begin{equation}
\label{eq:cost_total}
    \mathcal{E}_{fit}=\mathcal{E}_{data}+\mathcal{E}_{reg},
\end{equation}
\begin{equation}
\label{eq:cost_data}
    \mathcal{E}_{data}=\frac{1}{N}\sum_{n=1}^{N}\|M_{sub,n}^{fit}-M_{n}\|^{2}_{2},
\end{equation}
\begin{equation}
\label{eq:cost_reg}
    \mathcal{E}_{reg}=\lambda_{z}\mathcal{E}_{z}+\lambda_{\beta}\mathcal{E}_{\beta}+\lambda_{w}\mathcal{E}_{\theta_{w}}+\lambda_{\alpha}\mathcal{E}_{\alpha},
\end{equation}
where $M^{fit}_{sub,n}$ and $\theta_{w}\in\mathbb{R}^{6}$ denote the $n$-th row vector of $M^{fit}_{sub}$ and the axis-angle representation of both wrist joints. $\mathcal{E}_{z}$, $\mathcal{E}_{\beta}$, and $\mathcal{E}_{\theta_{w}}$ are the L2 regularization terms for $z$, $\beta$, and $\theta_{w}$, respectively. And $\mathcal{E}_{\alpha}$ is the exponential regularization term for preventing unnatural bending of the elbows and knees~\cite{SMPLify_Bogo_ECCV2016, SMPLify_X_2019_Pavlakos}. Each $\lambda$ represents regularization weight.

Joint coordinates $J=GM^{fit}\in\mathbb{R}^{17\times{}3}$ can be obtained from the fitted mesh $M^{fit}$ using a pretrained joint regression matrix $G\in\mathbb{R}^{17\times{}6890}$. The obtained $J$ is used to evaluate the joint coordinate estimation performance.

\section{Experimental Results}
\label{sec:experimental_results}

\subsection{Implementation Details}

The spatial sizes of the input image $I_c$ and 2D feature $F_{c}^{2D}$ are set to $(H_{0},W_{0})=(384,384)$ and $(H,W)=(96,96)$, respectively. The bounding box provided in the datasets is used to crop the human region from the input image. Random rotation is applied to the cuboid~\cite{2019_LT} along the vertical axis of the ground, and other augmentation is not used. Except for the fitting module, our network is trained end-to-end. Learnable parameters are included in the backbone and vertex regression module, and their learning rates are set to 1e-4 and 1e-3, respectively. The mini-batch size, number of epochs, number of sub-vertices $N$, and channel of the feature map $K$ is set to 3, 15, 108, and 32, respectively. The Adam optimizer~\cite{2015_Kingma} is used to train our network, which takes about 3.5 days using a single RTX 3090 GPU. Mesh sub-sampling algorithms~\cite{COMA_ECCV18} are applied to ground-truth human mesh vertices to obtain sub-sampled vertices, which are used for network training. The Adam optimizer is also used to update the optimization parameter $\Theta$ in the fitting module. The fitting module learning rate, number of iterations for fitting, $\lambda_{w}$, $\lambda_{z}$, $\lambda_{\beta}$, and $\lambda_{\alpha}$ are set to 6e-2, 500, 6e-2, 2e-6, 5e-6, and 5e-5, respectively. All regularization weights are simply determined through greedy search.

\subsection{Datasets}

Human3.6M~\cite{2014_H36M} is a large-scale dataset for 3D human pose estimation, including 3.6M video frames and 3D body joint annotations acquired from four synchronized cameras. It includes 11 human subjects (five females and six males), and according to previous works~\cite{2019_LT, 2020_Shin}, S1, S5, S6, S7, and S8 are used for training and S9, and S11 for testing. The SMPL mesh obtained by applying MoSh~\cite{mosh_Loper_2014} to Human3.6M is used for training and testing as ground-truth. The input image is undistorted before training and testing.

MPI-INF-3DHP~\cite{2017_3DHP} is a dataset for 3D human pose estimation and is obtained through the multi-camera markerless MoCap system. Since its test data includes single-view images, only train data composed of multi-view (i.e., 14) images are used in our experiments. Train data includes eight subjects. For a fair comparison, according to previous work~\cite{2020_Shin}, S1-S7 are used for training, S8 is used for testing, and views 0, 2, 7, and 8 are used among all cameras. MPI-INF-3DHP provides ground-truth 3D human joints, but does not provide ground-truth 3D human meshes, so pseudo ground-truth meshes are used to train the model. The pseudo ground-truth SMPL parameters are obtained by fitting the SMPL model to ground-truth 3D joints~\cite{SMPLify_X_2019_Pavlakos}, but the pseudo parameters are not used for evaluation.% because they do not provide accurate human shape and joint rotation information.

\subsection{Evaluation Metrics}

Mean-per-joint-position-error (MPJPE) is a metric that evaluates the performance of 3D human pose estimation based on the L2 distance between the predicted and ground-truth body joints. For LMT, joint coordinates in the world coordinate system can be estimated. Thus, following existing works~\cite{2019_LT, 2020_He_ET}, the L2 distance between the two joint sets is computed without aligning the predicted and ground-truth pelvis joints~\cite{2020_I2L, 2021_Lin_CVPR, 2021_Lin_ICCV, Choi_2020_ECCV_Pose2Mesh, 2018_Kanazawa, 2019_Kolotouros_CVPR, 2019_Kolotouros_ICCV}.

Mean-per-vertex-error (MPVE) is a metric that evaluates the performance of human mesh reconstruction based on the L2 distance between predicted and ground-truth mesh vertices. The proposed method is evaluated through MPVE only for the Human3.6M dataset on which ground-truth human meshes are available.

MPJPE and MPVE are used for the evaluation of human mesh reconstruction methods in most existing works. However, because MPJPE and MPVE measure the position errors for joints and vertices, they do not provide information on whether the rotation of the body part is accurately estimated. Therefore, the angular distance $d_{ang}$~\cite{Hartley_IJCV2013} is used between the estimated and ground-truth joint rotations for evaluating the proposed method:
\begin{equation}
\label{eq:angular_distance}
    {d}_{ang}=2\sin^{-1}{\frac{\|R-R^{*}\|_{F}}{2\sqrt{2}}},
\end{equation}
where, $R$, $R^{*}$, and $\|{\cdot}\|_F$ denote the predicted rotation matrix, ground-truth rotation matrix, and Frobenius norm, respectively. The joint rotation is defined relative to its parent joint. The rotation of the root joint (i.e., pelvis) denotes the global orientation of the entire body. All angular distances described in this paper are in degree units.

MPJPE averages the 3D position errors of all joints, so it cannot provide information about the case where only a specific joint has a large error. Therefore, 3DPCK~\cite{2017_3DHP} that computes the proportion of 3D joints with errors below a certain threshold is used. The AUC~\cite{2017_3DHP} is also presented for threshold-independent evaluation.

%%%%%%%%%%%%%%%%%%%%%%%%%%%%%%%%%%%%%%%%%%%%%%%%%%%%%%%%%%%%%%%%%%%%%%%%%%%%%%%%%%%%%%%%%%%%%%%%%%
\begin{table}[t]
\centering
{\small
\begin{tabular}{L{2.5cm}|C{1.2cm}|C{1.2cm}|C{1.4cm}}
\specialrule{.1em}{.05em}{.05em}
{Number of vertices} & {MPJPE} $\downarrow$ & {MPVE} $\downarrow$ & {Angular} $\downarrow$ \\ 
\hline
6890 & 19.85  & 25.21 & 11.98  \\ 
\cellcolor{Gray}431 & \cellcolor{Gray}18.40  & \cellcolor{Gray}\bf{24.15} & \cellcolor{Gray}11.60  \\ 
216 & 18.97  & 25.10 & 11.75  \\ 
\cellcolor{Gray}108 & \cellcolor{Gray}\bf{18.10}  & \cellcolor{Gray}24.88 & \cellcolor{Gray}\bf{11.54}  \\ 
54 & 18.35  & 26.47 & 12.00  \\ 
\specialrule{.1em}{.05em}{.05em}
\end{tabular}
}
\vspace*{-3mm}
\caption{\textbf{Ablation results for the number of estimated vertices on Human3.6M.} 3D heatmaps with $16\times{}16\times{}16$ resolution are used in all experiments in this table.}
\label{tab:ablation for sub-vertices model}
\vspace*{-1mm}
\end{table}
%%%%%%%%%%%%%%%%%%%%%%%%%%%%%%%%%%%%%%%%%%%%%%%%%%%%%%%%%%%%%%%%%%%%%%%%%%%%%%%%%%%%%%%%%%%%%%%%%%

%%%%%%%%%%%%%%%%%%%%%%%%%%%%%%%%%%%%%%%%%%%%%%%%%%%%%%%%%%%%%%%%%%%%%%%%%%%%%%%%%%%%%%%%%%%%%%%%%%
\begin{table}[t]
\centering
{\small
\begin{tabular}{L{2.5cm}|C{1.2cm}|C{1.2cm}|C{1.4cm}}
\specialrule{.1em}{.05em}{.05em}
{Heatmap resolution} & {MPJPE} $\downarrow$ & {MPVE} $\downarrow$ & {Angular} $\downarrow$ \\ 
\hline
$16\times{}16\times{}16$ & 18.10  & 24.88 & 11.54  \\ 
\cellcolor{Gray}$32\times{}32\times{}32$ & \cellcolor{Gray}18.02  & \cellcolor{Gray}24.19 & \cellcolor{Gray}11.45  \\ 
$64\times{}64\times{}64$ & \bf{17.59} & \bf{23.70} & \bf{11.33} \\ 
% \cellcolor{Gray}16R 108 & \cellcolor{Gray}\bf{18.10}  & \cellcolor{Gray}24.88 & \cellcolor{Gray}\bf{11.54}  \\ 
% 16R 54 & 18.35  & 26.47 & 12.00  \\ 
\specialrule{.1em}{.05em}{.05em}
\end{tabular}
}
\vspace*{-3mm}
\caption{\textbf{Ablation results for the resolution of 3D heatmaps on Human3.6M.} 108 vertices are estimated in all experiments in this table.}
\label{tab:ablation for heatmap resolution}
\vspace*{-3mm}
\end{table}
%%%%%%%%%%%%%%%%%%%%%%%%%%%%%%%%%%%%%%%%%%%%%%%%%%%%%%%%%%%%%%%%%%%%%%%%%%%%%%%%%%%%%%%%%%%%%%%%%%

\subsection{Ablation Experiments}

{\bf The number of sub-vertices.} The main problem of 3D heatmap-based prediction for SMPL mesh vertices is excessive GPU memory allocation for 3D heatmaps. This problem can be solved by estimating fewer sub-vertices. For example, if 108 sub-vertices are used instead of 6890 SMPL vertices, the size of GPU memory for the 3D heatmap is reduced by about $6890/108\approx{}63.8$ times.

To investigate the effect of using sub-vertices on mesh reconstruction performance, MPJPE, MPVE, and angular distance results are presented according to the number of vertices in Table~\ref{tab:ablation for sub-vertices model}. To compare the full-vertices model and all sub-vertices models under the same condition, $16\times{}16\times{}16$ heatmap resolution is used. It is the maximum resolution at which a full-vertices model under our computing resources can be trained.
%However, this may lead to deterioration of mesh reconstruction performance.

%In the case of estimating 6890 SMPL vertices using $64\times{}64\times{}64$ resolution 3D heatmaps, the size of GPU memory by the heatmap generated in forward pass is $64\times{}64\times{}64\times{}6890\times{}4 byte=6.89 GB$. However, under the same condition, when 108 sub-vertices are estimated instead of full-vertices, the size of memory by the heatmap is $64\times{}64\times{}64\times{}108\times{}4byte=0.11GB$, which is about 63.8 times smaller than the case of full-vertices. 

Table~\ref{tab:ablation for sub-vertices model} shows that better quantitative results are obtained in most cases using sub-vertices than when using full-vertices. Only for the 54 sub-vertices, MPVE and angular distance performances deteriorate compared to full-vertices. This degraded performance is due to the fact that 54 sub-vertices do not provide sufficient information for joint rotation and shape reconstruction, given the supplementary material. We adopt the 108 vertices model that shows the best MPJPE and angular distance performance and requires a relatively smaller heatmap size.

{\bf Heatmap resolution.} Experiments using various heatmap resolutions are conducted to investigate a model that can accurately estimate 108 sub-vertices. Table~\ref{tab:ablation for heatmap resolution} shows the performance for the cases in which the heatmap resolution is set to $16\times{}16\times{}16$, $32\times{}32\times{}32$, and $64\times{}64\times{}64$. The proposed method shows the best performance at $64\times{}64\times{}64$ heatmap resolution. In this case, the memory allocation for the heatmap is $64\times{}64\times{}64\times{}108\times{}4byte=113.2MB$, which is similar to the memory allocation of $16\times{}16\times{}16\times{}6890\times{}4byte=112.9MB$ for the maximum heatmap resolution allowed by full-vertices. According to the performance comparison of the full-vertices model in Table~\ref{tab:ablation for sub-vertices model} and the sub-vertices model in Table~\ref{tab:ablation for heatmap resolution}, the 108 vertices model with $64\times{}64\times{}64$ heatmap resolution achieves better performance for all evaluation metrics than the full-vertices model without additional memory cost.

\begin{figure}[t]
\centering
\includegraphics[width=\linewidth]{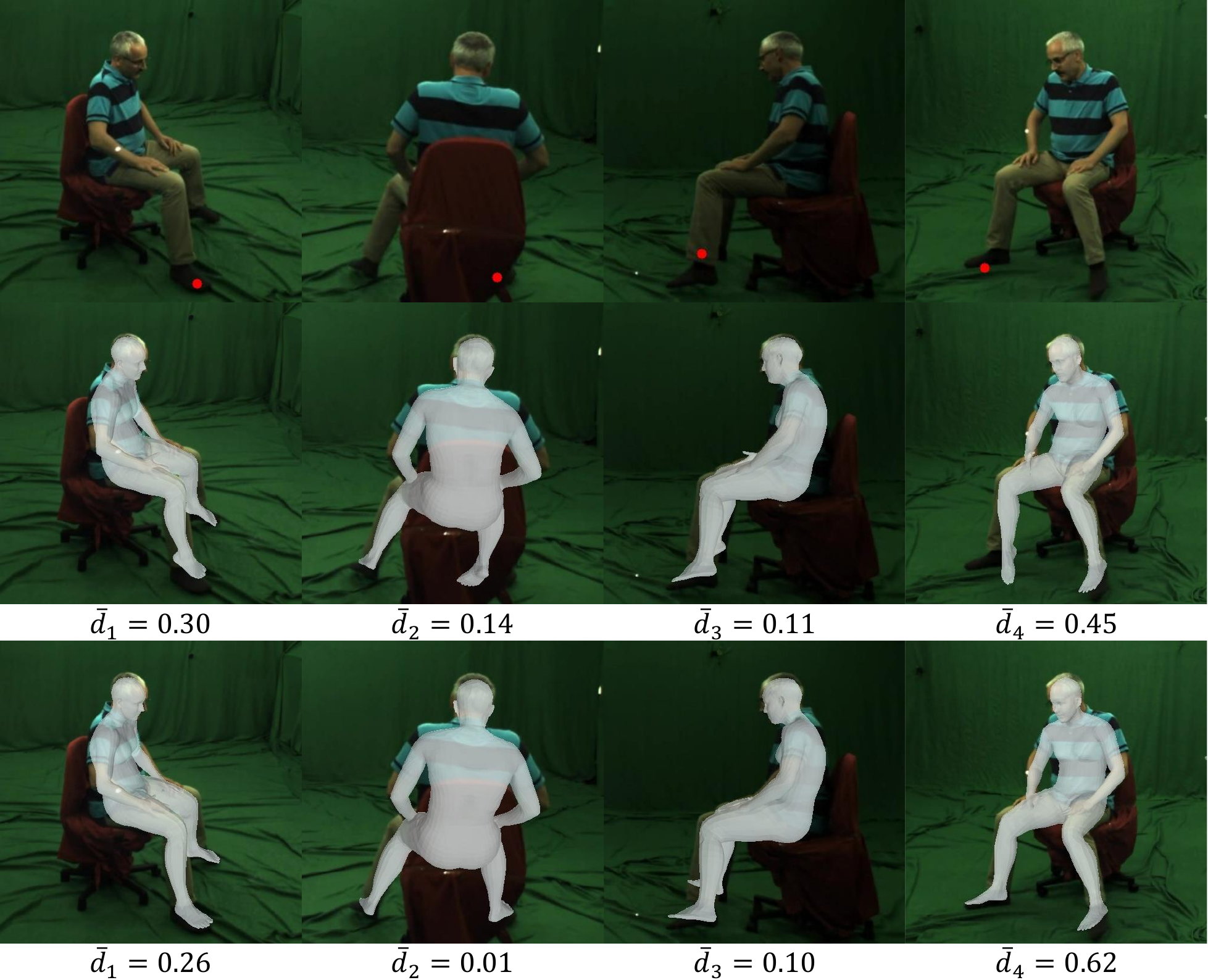}
\vspace*{-4mm}
\caption{The first row visualizes the input multi-view images. The second and third rows show the reconstructed meshes generated from the softmax baseline and LMT, respectively. $\Bar{d}_c$ is obtained by averaging the confidence weight $d_c$ corresponding to the red pixel on the multi-view image along the channel axis. The red pixels are obtained by projecting the voxel including a ground-truth vertex on the right foot to each image plane. Therefore, $\Bar{d}_c$ indicates how much the model depends on the image feature obtained from each view $c$ to construct the aggregated feature of the voxel containing the right foot vertex.}
\label{fig3}
\vspace*{-4mm}
\end{figure}

{\bf Multi-view inconsistency.} To prove that using visibility helps feature aggregation, we implement the \emph{softmax baseline} that does not use visibility, and compare it to the LMT. The Softmax baseline generates the image feature $F^{2D}_{c}$ by directly feeding $F^{deconv}_{c}$ into the $1\times{}1$ convolution layer without concatenation with visibility $v_{c}$. When aggregating multi-view features, using only features obtained from views in which the human body surface is visible is desirable, because this leads to the consistency of aggregated multi-view features. However, in the softmax baseline, the multi-view inconsistency problem may arise, which can be mitigated through the use of visibility.

In the first and fourth views of Fig.~\ref{fig3}, the right foot is clearly visible. In the third view, the right foot is not visible, but it can be contextually inferred that it is behind the left foot. However, in the second view, severe occlusion prevents the right foot from being estimated. Therefore, relying on the features obtained from the remaining views rather than the second view is preferable for estimating the position of the right foot. However, the softmax baseline shows a higher dependence on the second view than on the third view, which causes the model to incorrectly estimate the right foot mesh. On the other hand, LMT uses visibility to reduce the dependence on the second view and increase the dependence on the remaining views. Consequently, LMT successfully reconstructs the right foot mesh.

%%%%%%%%%%%%%%%%%%%%%%%%%%%%%%%%%%%%%%%%%%%%%%%%%%%%%%%%%%%%%%%%%%%%%%%%%%%%%%%%%%%%%%%%%%%%%%%%%%
\begin{table}[t]
\centering
{\small
\begin{tabular}{L{1.8cm}|C{1.1cm}|C{1.1cm}|C{1.1cm}|C{1.1cm}}
\specialrule{.1em}{.05em}{.05em}
{Model} & {MPVE} $\dagger$ & {MPJPE} & {MPVE} & {Angular} \\ 
\hline
Softmax & 22.12 & 17.84 & 24.14 & 11.42 \\
LMT & \bf{21.50} & \bf{17.59} & \bf{23.70} & \bf{11.33}  \\ 
\specialrule{.1em}{.05em}{.05em}
\end{tabular}
}
\vspace*{-3mm}
\caption{\textbf{Comparison with the softmax baseline on Human3.6M.} $\dagger$ means that the regressed vertices $M$ from the vertex regression module are evaluated.}
\label{tab:comparison with softmax on human36m test data}
\vspace*{-1mm}
\end{table}
%%%%%%%%%%%%%%%%%%%%%%%%%%%%%%%%%%%%%%%%%%%%%%%%%%%%%%%%%%%%%%%%%%%%%%%%%%%%%%%%%%%%%%%%%%%%%%%%%%

%%%%%%%%%%%%%%%%%%%%%%%%%%%%%%%%%%%%%%%%%%%%%%%%%%%%%%%%%%%%%%%%%%%%%%%%%%%%%%%%%%%%%%%%%%
% MPI-INF-3DHP generalization test
\begin{table}[t]
\scriptsize
% \footnotesize
\centering
\setlength\tabcolsep{1.0pt}
\def\arraystretch{1.1}
\begin{tabular}{L{1.3cm}|C{0.70cm}C{0.70cm}C{0.70cm}C{0.70cm}C{0.70cm}C{0.70cm}C{0.70cm}C{0.70cm}C{0.70cm}}
% \specialrule{.1em}{.05em}{.05em}
% \begin{tabular}{L{1.3cm}|C{0.70cm}C{0.70cm}C{0.70cm}C{0.70cm}C{0.70cm}C{0.70cm}C{0.70cm}C{0.70cm}C{0.70cm}}
\specialrule{.1em}{.05em}{.05em}
Model & S1 & S2 & S3 & S4 & S5 & S6 & S7 & S8 & \bf{Avg} \\ \hline
\multicolumn{1}{l}{\textbf{\textit{MPJPE}} $\downarrow$}  \\ \hline
Softmax & 85.83 & 64.03 & 63.42 & 82.55 & 65.18 & 71.68 & 66.02 & 70.74 & 70.66 \\
LMT & \bf{81.32} & \bf{61.49} & \bf{60.50} & \bf{78.62} & \bf{62.47} & \bf{71.39} & \bf{65.77} & \bf{66.78} & \bf{68.02} \\
\hline
\hline
\multicolumn{1}{l}{\textbf{\textit{3DPCK}} $\uparrow$}  \\ \hline
Softmax & 89.16 & 96.68 & 95.01 & 88.67 & 94.37 & 93.42 & 94.05 & 94.04 & 93.32 \\
LMT & \bf{90.28} & \bf{97.80} & \bf{95.85} & \bf{89.81} & \bf{95.21} & \bf{93.52} & \bf{94.91} & \bf{95.22} & \bf{94.07} \\
\hline
\hline
\multicolumn{1}{l}{\textbf{\textit{AUC}} $\uparrow$}  \\ \hline
Softmax & 53.00 & 59.97 & 62.10 & 58.01 & 60.11 & 58.06 & 60.44 & 57.53 & 58.91 \\
LMT & \bf{53.94} & \bf{60.47} & \bf{62.54} & \bf{58.66} & \bf{60.32} & \bf{58.12} & \bf{60.53} & \bf{58.33} & \bf{59.30} \\
\specialrule{.1em}{.05em}{.05em}
\end{tabular}
\vspace*{-3mm}
\caption{\textbf{Cross-dataset evaluation of the softmax baseline and LMT.} The two models are trained on Human3.6m and evaluated on MPI-INF-3DHP. S1-S8 denote the subjects in MPI-INF-3DHP.}
\vspace*{-1mm}
\label{table:MPII3D_generalization_test}
\end{table}
%%%%%%%%%%%%%%%%%%%%%%%%%%%%%%%%%%%%%%%%%%%%%%%%%%%%%%%%%%%%%%%%%%%%%%%%%%%%%%%%%%%%%%%%%%

%%%%%%%%%%%%%%%%%%%%%%%%%%%%%%%%%%%%%%%%%%%%%%%%%%%%%%%%%%%%%%%%%%%%%%%%%%%%%%%%%%%%%%%%%%
\begin{table}[t]
\centering
{\scriptsize
\begin{tabular}{L{2.7cm}|C{1.3cm}|C{1.3cm}|C{1.3cm}}
\specialrule{.1em}{.05em}{.05em}
{Model} & {MPJPE} $\downarrow$ & {MPVE} $\downarrow$ & {Angular} $\downarrow$ \\ 
\hline
LT-fitting~\cite{2019_LT,lightcap2021} & \bf{16.21}  & 35.20 & 15.73  \\ 
\cellcolor{Gray}LT-fitting~\cite{2019_LT,lightcap2021} (w/o reg) & \cellcolor{Gray}16.40  & \cellcolor{Gray}42.99 & \cellcolor{Gray}22.94  \\ 
LMT & 17.59  & \bf{23.70} & \bf{11.33}  \\ 
\cellcolor{Gray}LMT (w/o reg) & \cellcolor{Gray}17.48  & \cellcolor{Gray}25.30 & \cellcolor{Gray}13.03  \\ 
\specialrule{.1em}{.05em}{.05em}
\end{tabular}
}
\vspace*{-3mm}
\caption{\textbf{Comparison with joint fitting on Human3.6M.} ``w/o reg'' means that no regularization term $\mathcal{E}_{reg}$ is used.}
\label{tab:comparison_joint-fitting_surface-fitting}
\vspace*{-3mm}
\end{table}
%%%%%%%%%%%%%%%%%%%%%%%%%%%%%%%%%%%%%%%%%%%%%%%%%%%%%%%%%%%%%%%%%%%%%%%%%%%%%%%%%%%%%%%%%%

%%%%%%%%%%%%%%%%%%%%%%%%%%%%%%%%%%%%%%%%%%%%%%%%%%%%%%%%%%%%%%%%%%%%%%%%%%%%%%%%%%%%%%%%%%%%%%%%%%
\begin{table*}[t]
\scriptsize
%\footnotesize
\centering
\setlength\tabcolsep{1.0pt}
\def\arraystretch{1.1}
\begin{tabular}{L{2.0cm}|C{0.63cm}C{0.60cm}C{0.60cm}C{0.60cm}C{0.70cm}C{0.75cm}C{0.60cm}C{0.70cm}C{0.70cm}C{0.60cm}C{0.60cm}C{0.70cm}C{0.70cm}C{0.60cm}C{0.70cm}C{0.70cm}C{0.75cm}C{0.75cm}C{0.70cm}C{0.70cm}}
\specialrule{.1em}{.05em}{.05em}
Angular $\downarrow$ & pelvis & L-hip & R-hip & torso & L-knee & R-knee & spine & L-ankl & R-ankl & chest & neck & L-thrx & R-thrx & head & L-shld & R-shld & L-elbw & R-elbw & L-wrst & R-wrst \\ \hline
LT-fitting~\cite{2019_LT,lightcap2021} & 8.18 & 10.10 & 9.37 & 10.75 & 9.17 & 9.21 & 7.8 & 17.31 & 16.86 & 5.88 & \bf{12.07} & 10.72 & 11.64 & 12.52 & \bf{11.65} & 14.18 & 20.24 & 16.14 & 43.00 & 43.20 \\
LMT & \bf{4.77} & \bf{5.69} & \bf{5.79} & \bf{6.40} & \bf{5.80} & \bf{5.38} & \bf{5.68} & \bf{8.58} & \bf{9.85} & \bf{4.48} & 12.32 & \bf{9.39} & \bf{10.22} & \bf{10.69} & 11.86 & \bf{14.06} & \bf{13.45} & \bf{11.50} & \bf{19.53} & \bf{20.22}  \\
\specialrule{.1em}{.05em}{.05em}
\end{tabular}
\vspace*{-3mm}
\caption{\textbf{Per-joint rotation error comparison with joint fitting on Human3.6M.}}
\vspace*{-1mm}
\label{table:detailed_angular_distance_joint-fitting_surface-fitting}
\end{table*}
%%%%%%%%%%%%%%%%%%%%%%%%%%%%%%%%%%%%%%%%%%%%%%%%%%%%%%%%%%%%%%%%%%%%%%%%%%%%%%%%%%%%%%%%%%%%%%%%%%

%%%%%%%%%%%%%%%%%%%%%%%%%%%%%%%%%%%%%%%%%%%%%%%%%%%%%%%%%%%%%%%%%%%%%%%%%%%%%%%%%%%%%%%%%%
% SOTA, Human3.6M
% \begin{table}[t]
% \centering
% {\footnotesize
% \begin{tabular}{L{1.8cm}|C{0.65cm}|C{0.65cm}|C{0.9cm}|C{0.8cm}|C{1.0cm}}
% \specialrule{.1em}{.05em}{.05em}
% {Model} & {Net} & {Scale} & {MPJPE} & {MPVE} & {Angular} \\ 
% \hline
% \scriptsize{LT-fitting~\cite{2019_LT,lightcap2021}} & R152 & 384 & \bf{16.21}  & 35.20 & 15.73  \\
% \cellcolor{Gray}LMT & \cellcolor{Gray}R152 & \cellcolor{Gray}384 & \cellcolor{Gray}17.59  & \cellcolor{Gray}\bf{23.70} & \cellcolor{Gray}\bf{11.33}  \\ 
% \hline
% \hline
% \cite{2020_Shin} & R50 & 224 & 46.9  & - & -  \\ 
% \cellcolor{Gray}LMT & \cellcolor{Gray}R50 & \cellcolor{Gray}224 & \cellcolor{Gray}\bf{30.56}  & \cellcolor{Gray}42.28 & \cellcolor{Gray}14.61  \\ 
% \specialrule{.1em}{.05em}{.05em}
% \end{tabular}
% }
% \vspace*{-3mm}
% \caption{\textbf{The results of evaluation on the Human3.6M.}}
% \label{tab:SOTA_on_human36m}
% \vspace*{-3mm}
% \end{table}

% SOTA, Human3.6M
\begin{table}[t]
\scriptsize
\centering
\begin{tabular}{L{3.3cm}|C{1.1cm}|C{1.1cm}|C{1.1cm}}
\specialrule{.1em}{.05em}{.05em}
{Model} & {MPJPE $\downarrow$} & {MPVE $\downarrow$} & {Angular $\downarrow$} \\ 
\hline
(R50-224) Parameter regr.~\cite{2020_Shin} & 46.90 & - & -  \\ 
\cellcolor{Gray}(R50-224) LMT & \cellcolor{Gray}\bf{30.56}  & \cellcolor{Gray}42.28 & \cellcolor{Gray}14.61  \\ 
\hline
\hline
(R152-384) LT-fitting~\cite{2019_LT,lightcap2021} & \bf{16.21}  & 35.20 & 15.73  \\
\cellcolor{Gray}(R152-384) LMT & \cellcolor{Gray}17.59  & \cellcolor{Gray}\bf{23.70} & \cellcolor{Gray}\bf{11.33}  \\ 
\specialrule{.1em}{.05em}{.05em}
\end{tabular}
\vspace*{-3mm}
\caption{\textbf{Comparison results on Human3.6M.} ``R50-224'' means that ResNet-50 backbone and input image of $224\times{224}$ resolution are used. Similarly, ``R152-384'' denotes ResNet-152 and $384\times{384}$ resolution.}
\label{tab:SOTA_on_human36m}
\vspace*{-1mm}
\end{table}

% % SOTA, Human3.6M
% \begin{table}[t]
% \centering
% {\footnotesize
% \begin{tabular}{L{3.4cm}|C{1.1cm}|C{1.0cm}|C{1.2cm}}
% \specialrule{.1em}{.05em}{.05em}
% {Model} & {MPJPE $\downarrow$} & {MPVE $\downarrow$} & {Angular $\downarrow$} \\ 
% \hline
% \scriptsize{R152-384$\times{}$384 LT-fitting~\cite{2019_LT,lightcap2021}} & \bf{16.21}  & 35.20 & 15.73  \\
% \cellcolor{Gray}R152-384$\times{}$384 LMT & \cellcolor{Gray}17.59  & \cellcolor{Gray}23.70 & \cellcolor{Gray}11.33  \\ 
% \hline
% \hline
% R50-224$\times{}$224 \cite{2020_Shin} & 46.9  & - & -  \\ 
% \cellcolor{Gray}R50-224$\times{}$224 LMT & \cellcolor{Gray}\bf{30.56}  & \cellcolor{Gray}42.28 & \cellcolor{Gray}14.61  \\ 
% \specialrule{.1em}{.05em}{.05em}
% \end{tabular}
% }
% \vspace*{-3mm}
% \caption{\textbf{The results of evaluation on the Human3.6M.}}
% \label{tab:SOTA_on_human36m}
% \vspace*{-3mm}
% \end{table}
%%%%%%%%%%%%%%%%%%%%%%%%%%%%%%%%%%%%%%%%%%%%%%%%%%%%%%%%%%%%%%%%%%%%%%%%%%%%%%%%%%%%%%%%%%

%%%%%%%%%%%%%%%%%%%%%%%%%%%%%%%%%%%%%%%%%%%%%%%%%%%%%%%%%%%%%%%%%%%%%%%%%%%%%%%%%%%%%%%%%%
% SOTA, MPI-INF-3DHP
\begin{table}[t]
\scriptsize
\centering
\begin{tabular}{L{3.3cm}|C{1.1cm}|C{1.1cm}|C{1.1cm}}
\specialrule{.1em}{.05em}{.05em}
{Model} & {MPJPE $\downarrow$} & {3DPCK $\uparrow$} & {AUC $\uparrow$} \\ 
\hline
(R50-224) Parameter regr.~\cite{2020_Shin} & 50.20  & \bf{97.40} & 65.60  \\ 
\cellcolor{Gray}(R50-224) LMT & \cellcolor{Gray}\bf{45.87}  & \cellcolor{Gray}96.59 & \cellcolor{Gray}\bf{71.57}  \\ 
\hline
\hline
(R152-384) LT-fitting~\cite{2019_LT,lightcap2021} & \bf{33.33}  & \bf{99.60} & \bf{77.23}  \\
\cellcolor{Gray}(R152-384) LMT & \cellcolor{Gray}33.70  & \cellcolor{Gray}99.37 & \cellcolor{Gray}77.09  \\ 
\specialrule{.1em}{.05em}{.05em}
\end{tabular}
\vspace*{-3mm}
\caption{\textbf{Comparison results on MPI-INF-3DHP.}}
\label{tab:SOTA_on_MPI-INF-3DHP}
\vspace*{-3mm}
\end{table}
%%%%%%%%%%%%%%%%%%%%%%%%%%%%%%%%%%%%%%%%%%%%%%%%%%%%%%%%%%%%%%%%%%%%%%%%%%%%%%%%%%%%%%%%%%

{\bf Effect of using visibility.} We investigate the quantitative results of using per-vertex visibility in terms of joint coordinates, rotations, and shape estimation. Table~\ref{tab:comparison with softmax on human36m test data} shows the results when the softmax baseline and LMT are trained on Human3.6M train data and evaluated on Human3.6M test data. The second column in Table~\ref{tab:comparison with softmax on human36m test data} shows the MPVE for sub-vertices estimated by the vertex regression module, which proves that using visibility helps the network to estimate the human surface accurately. Columns 3-5 of Table~\ref{tab:comparison with softmax on human36m test data} show that the use of visibility helps to improve MPJPE, MPVE, and angular distance results even after fitting.

{\bf Generalization.} The proposed method exploits geometry information (i.e., visibility) obtained from a single-view model for feature aggregation. This single-view model can be trained using more various datasets than the multi-view model. Therefore, the use of geometry information from the single-view model causes the effect of implicit learning through such various datasets and helps to improve the generalization performance of the proposed method. To prove this quantitatively, we train the softmax baseline and LMT using Human3.6M train data and evaluate them for all subjects of MPI-INF-3DHP. Despite the differences between the two datasets, Table~\ref{table:MPII3D_generalization_test} shows that the LMT significantly outperforms the softmax baseline in all metrics evaluating the joint coordinate estimation performance.

%All subjects in Human3.6M dataset wear clothes that reveal their body shape, but subjects in MPI-INF-3DHP dataset wear looser clothes compared to Human3.6M. Also, the chair used in the MPI-INF-3DHP dataset causes more serious occlusion than Human3.6M. 

{\bf Comparison with joint fitting.} We demonstrate that fitting on the human surface brings more benefits than fitting on the human joint~\cite{lightcap2021}. However, the method of ~\cite{lightcap2021} cannot be directly compared with LMT because it is for multi-person mesh reconstruction. Therefore, we design the \emph{LT-fitting baseline} using the state-of-the-art multi-view joint estimation method LT~\cite{2019_LT}. LT-fitting modifies $\mathcal{E}_{data}$ to minimize the difference between the predicted and ground-truth joints, and uses the same regularization term as LMT.

\begin{figure}[t]
\centering
\includegraphics[width=\linewidth]{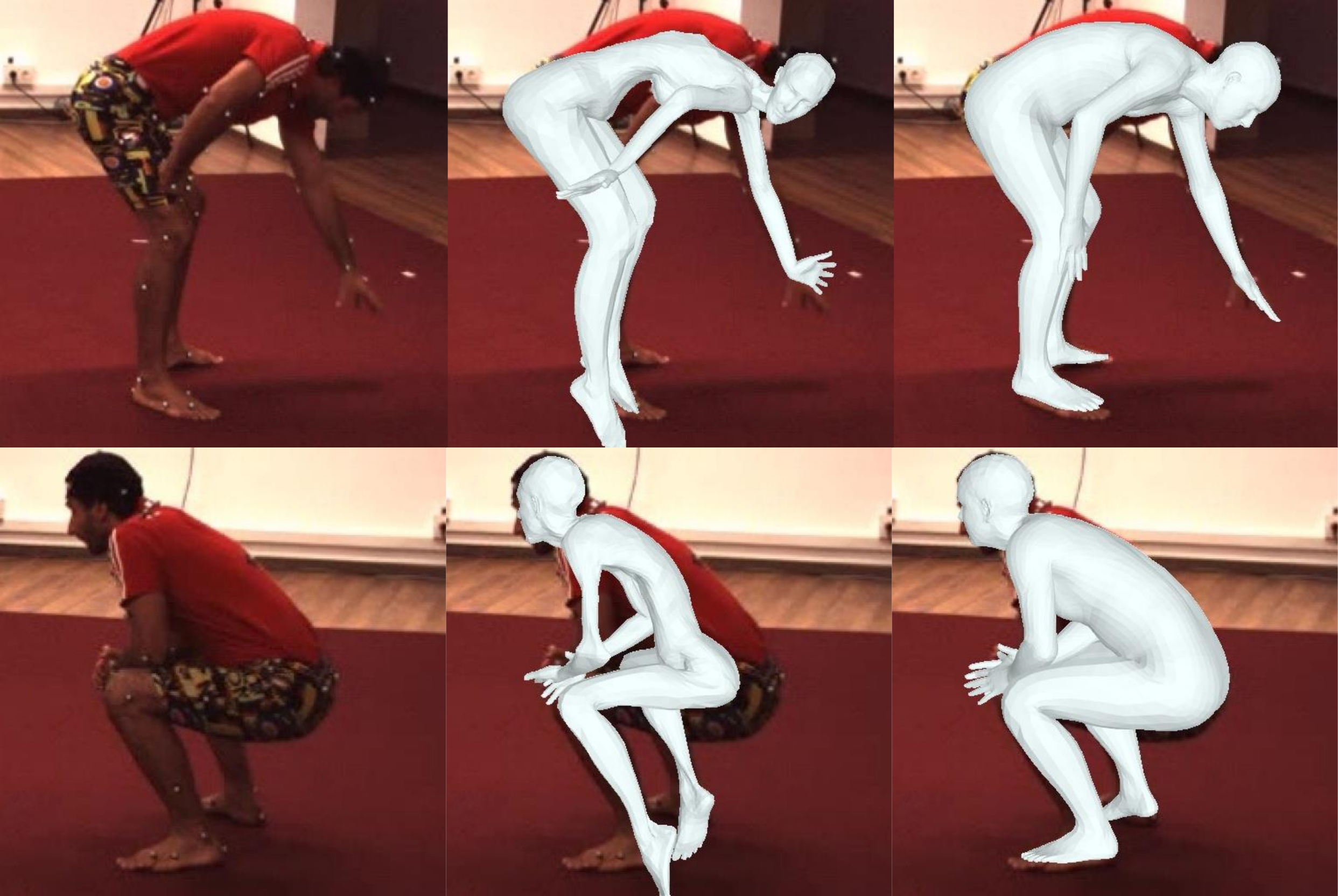}
\vspace*{-4mm}
\caption{\textbf{Qualitative comparison with joint fitting on Human3.6M.} The first column shows the input images. The second and third columns visualize the meshes reconstructed by LT-fitting and LMT, respectively.}
\label{fig4_shape_comparison}
\vspace*{-4mm}
\end{figure}

Table~\ref{tab:comparison_joint-fitting_surface-fitting} shows MPJPE, MPVE, and angular distance results of LT-fitting and LMT evaluated on Human3.6M. In both cases of LT-fitting and LMT, the use of regularization results in better joint rotation and shape estimation. And LT-fitting relies more heavily on regularization than LMT. However, LT-fitting with regularization shows worse MPVE and angular distance results than LMT without regularization. Table~\ref{table:detailed_angular_distance_joint-fitting_surface-fitting} shows the rotation errors of LT-fitting and LMT for each joint. For most joints, the rotation prediction performance of LMT is significantly better than that of LT-fitting. Fig.~\ref{fig4_shape_comparison} shows the human mesh reconstruction by LT-fitting and LMT. LT-fitting cannot describe the subject's body shape well because it cannot resolve the ambiguity of the human shape. On the other hand, LMT shows a visually satisfactory result. All these results show that using the human surface rather than the human joint is beneficial for human pose and shape estimation.

% extended joint fitting < surface fitting
% shape estimation
% These results show that it is beneficial to use human surface rather than human joint for human pose and shape estimation.

% In particular, for the wrist joint, LT-fitting shows more than twice the error of LMT. Fig.~\ref{fig1} qualitatively shows the superiority of surface fitting. In the case of joint fitting, the reconstructed mesh does not align well with the image because the joint rotation accuracy is low at the wrist and ankle. In contrast, our surface fitting method better reconstructs the human mesh. These results show that it is beneficial to use human surface rather than human joint for human pose and shape estimation.

\subsection{Comparison on Human3.6M}

Table~\ref{tab:SOTA_on_human36m} shows the results of previous multi-view human mesh reconstruction methods and LMT trained and evaluated on Human3.6M. The same input image size and the same backbone are used for a fair comparison with the \emph{parameter regression} method of \cite{2020_Shin}. Since \cite{2020_Shin} does not provide MPVE and angular distance results, MPJPE is used for comparison, which shows that LMT significantly outperforms the method of \cite{2020_Shin}. These results show that the combination of heatmap-based vertex regression and subsequent SMPL fitting brings more accurate results than the method of directly regressing the SMPL parameters from input images. LT-fitting is different from LMT in that SMPL is fitted to the human joint rather than the human surface. Due to this difference, LT-fitting does not obtain enough information to resolve the ambiguity for joint rotation and human shape determination, and as a result achieves significantly lower MPVE and angular distance performance than LMT.

\begin{figure}[t]
\centering
\includegraphics[width=\linewidth]{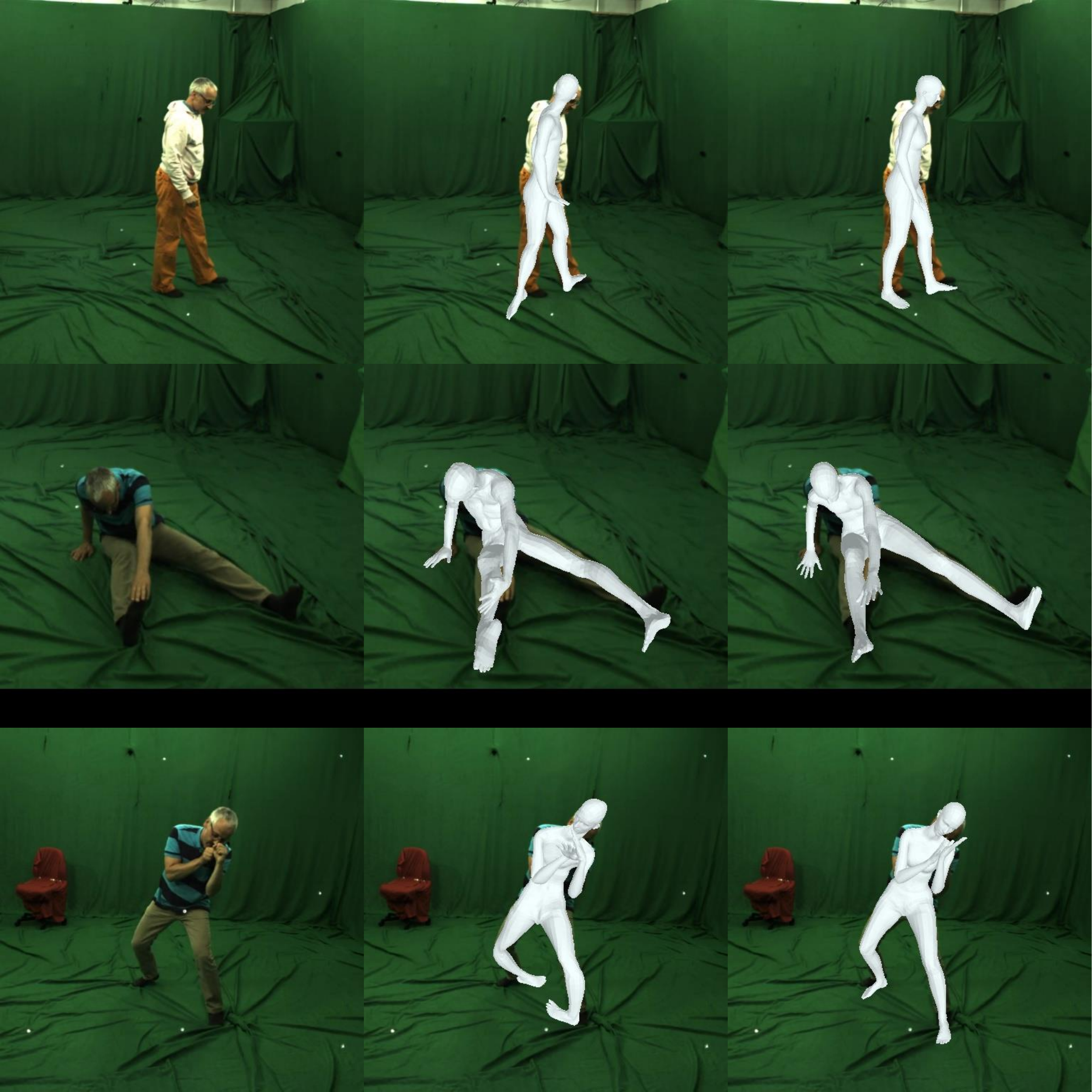}
\vspace*{-4mm}
\caption{\textbf{Qualitative comparison with joint fitting on MPI-INF-3DHP.} The first column shows the input images. The second and third columns visualize the meshes reconstructed by LT-fitting and LMT, respectively.}
\label{fig:qualitative results on MPI-INF-3DHP}
\vspace*{-4mm}
\end{figure}

\subsection{Comparison on MPI-INF-3DHP}

Table~\ref{tab:SOTA_on_MPI-INF-3DHP} shows the results of previous multi-view human mesh reconstruction methods and LMT trained and evaluated on MPI-INF-3DHP. For a fair comparison with \cite{2020_Shin}, the LMT model is pretrained on Human3.6M and then fine-tuned on MPI-INF-3DHP. For 3DPCK with a threshold of 150 mm, \cite{2020_Shin} shows better results than LMT, but for threshold-independent AUC, LMT shows better results. Also, as in Human3.6M, LMT shows better MPJPE performance. The LMT model shows a competitive joint coordinate estimation result with LT-fitting. In the case of MPI-INF-3DHP, ground-truth SMPL parameters are not provided, so joint rotation and shape estimation results are not presented. However, LMT gives qualitatively better mesh reconstruction results than LT-fitting, as shown in Fig.~\ref{fig:qualitative results on MPI-INF-3DHP}.
% given the supplementary material

%-------------------------------------------------------------------------
\section{Conclusion}
\label{sec:conclusion}

%In this paper, a heatmap-based method to reconstruct a single human mesh from multi-view images is proposed.
In this paper, a two-stage method consisting of visibility-based sub-vertices estimation and surface fitting is proposed to reconstruct a single human mesh from multi-view images. The estimation of sub-vertices rather than full-vertices solves the problem of excessive GPU memory usage. In addition, the use of per-vertex visibility improves the mesh vertices estimation performance by alleviating the multi-view inconsistency problem. Surface fitting is also demonstrated to help estimate joint rotations and human shape compared to joint fitting. According to the experimental results, the proposed LMT significantly outperforms the existing multi-view human mesh reconstruction methods on the Human3.6M and MPI-INF-3DHP datasets. However, since using a single-view mesh reconstruction model to acquire visibility complicates the proposed model, additional studies are needed for a more efficient method to obtain visibility information. In addition, the investigation of more diverse viewpoints and in-the-wild input images is another future work.

%%%%%%%%%%%%%%%%%%%%%%%%%%%%%%%
\section*{Supplementary Material}

In the supplementary material, we first provide the detailed architecture of the vertex regression module. We then give the additional results of the 54 vertices model, single-view method in the visibility module, and multi-view inconsistency. Also, we present qualitative comparison of surface fitting and joint fitting methods.
% In the supplementary material, we provide the detailed architecture of the vertex regression module, and the location of sub-vertices. Also, we present additional ablation study on 54 vertices model, single-view method constituting the visibility module, multi-view inconsistency, and qualitative comparison of surface fitting and joint fitting methods.

\renewcommand{\thesection}{S\arabic{section}}
\renewcommand{\thetable}{S\arabic{table}}
\renewcommand{\thefigure}{S\arabic{figure}}
\setcounter{section}{0}
\setcounter{table}{0}
\setcounter{figure}{0}

%-------------------------------------------------------------------------
\section{Vertex Regression Module}

\begin{figure*}[t]
\centering
\includegraphics[width=\linewidth]{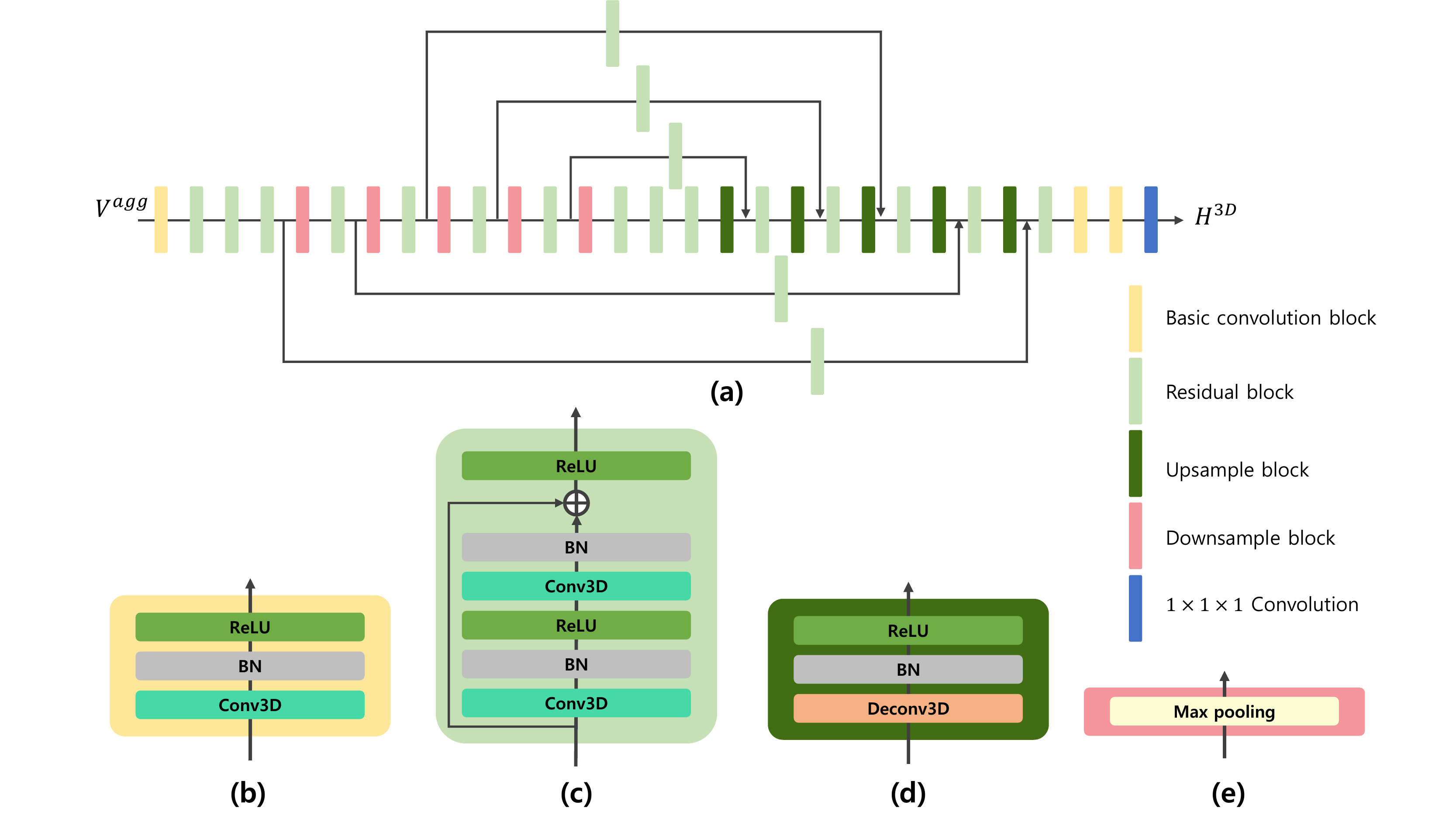}
\vspace*{-4mm}
\caption{\textbf{The architecture of the vertex regression module.} (a) Pipeline of the vertex regression module. (b) Basic convolution block. (c) Residual block. (d) Upsample block. (e) Downsample block.}
\label{fig:vertex regression module}
\vspace*{-1mm}
\end{figure*}

The vertex regression module consists of basic convolution blocks, residual blocks, downsample blocks, upsample blocks, and a $1\times{}1\times{}1$ convolution layer. The basic convolution block consists of a 3D convolution layer, a batch normalization layer, and a ReLU activation function. The residual block contains two 3D convolution layers, two batch normalization layers, two ReLU activation functions, and a residual connection. The downsample block consists of a 3D max pooling layer with a stride of 2. The upsample block consists of a 3D deconvolution layer with a stride of 2, a batch normalization layer, and a ReLU activation function. The vertex regression module is constructed using 3 basic convolution blocks, 20 residual blocks, 5 downsample blocks, 5 upsample blocks, and a $1\times{}1\times{}1$ convolution. Fig.~\ref{fig:vertex regression module} shows the detailed structure of the vertex regression module.

\begin{figure*}[t]
\centering
\includegraphics[width=\linewidth]{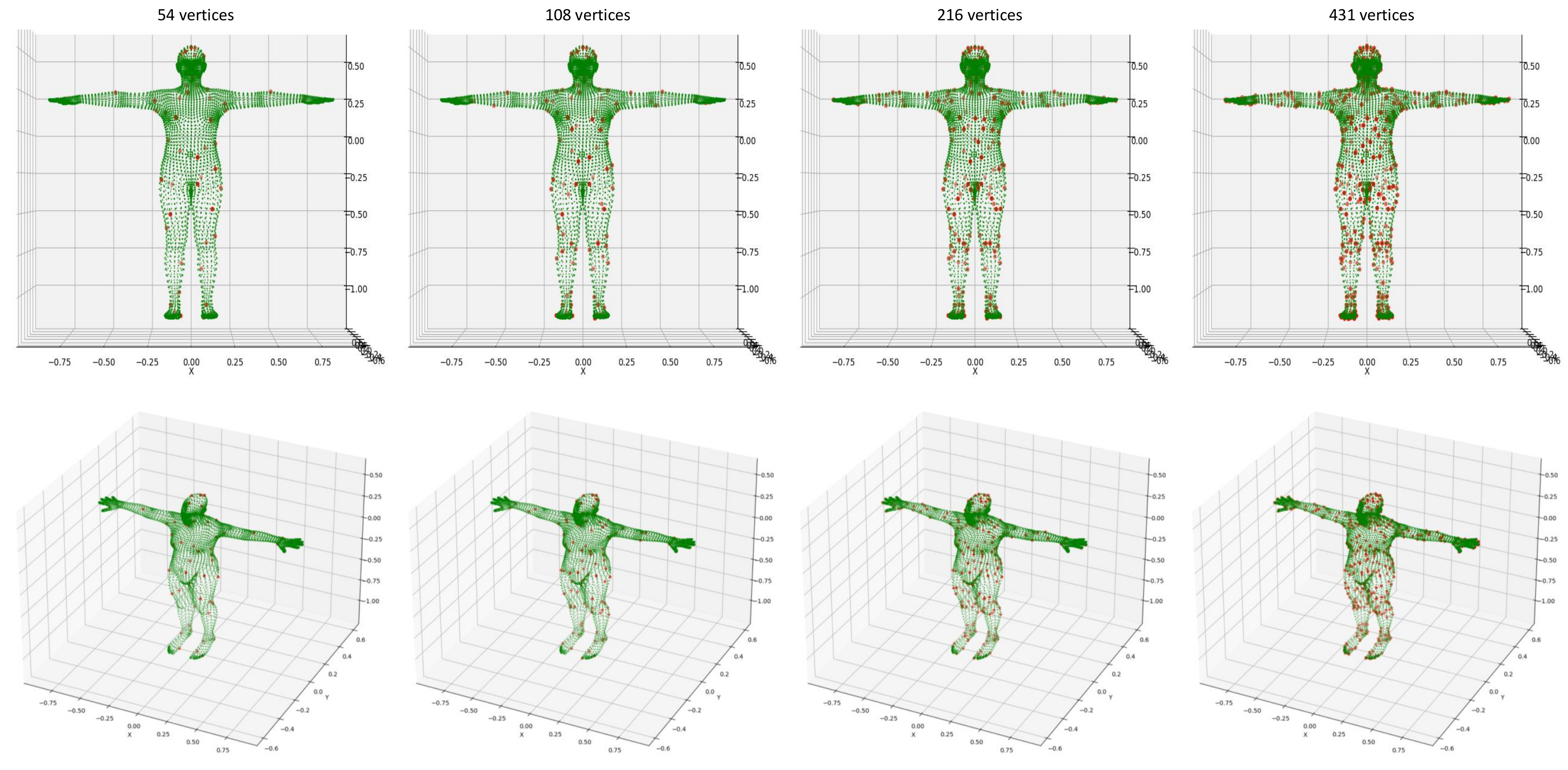}
\vspace*{-1mm}
\caption{\textbf{Visualization of the vertex positions of sub-vertices models.} Green and red point sets denote full-vertices and sub-vertices, respectively.}
\label{fig:sub-vertices location}
\vspace*{-4mm}
\end{figure*}

%-------------------------------------------------------------------------
\section{54 Vertices Model}

%%%%%%%%%%%%%%%%%%%%%%%%%%%%%%%%%%%%%%%%%%%%%%%%%%%%%%%%%%%%%%%%%%%%%%%%%%%%%%%%%%%%%%%%%%%%%%%%%%
% 108 vertices vs 54 vertices on 16x16x16 heatmap resolution
\begin{table*}[t]
\scriptsize
%\footnotesize
\centering
\setlength\tabcolsep{1.0pt}
\def\arraystretch{1.1}
\begin{tabular}{L{1.7cm}|C{0.63cm}C{0.60cm}C{0.60cm}C{0.60cm}C{0.70cm}C{0.75cm}C{0.60cm}C{0.70cm}C{0.70cm}C{0.60cm}C{0.60cm}C{0.70cm}C{0.70cm}C{0.60cm}C{0.70cm}C{0.70cm}C{0.75cm}C{0.75cm}C{0.70cm}C{0.70cm}}
\specialrule{.1em}{.05em}{.05em}
Angular $\downarrow$ & pelvis & L-hip & R-hip & torso & L-knee & R-knee & spine & L-ankl & R-ankl & chest & neck & L-thrx & R-thrx & head & L-shld & R-shld & L-elbw & R-elbw & L-wrst & R-wrst \\ \hline
108 & 5.09 & \bf{5.75} & \bf{5.89} & \bf{5.80} & \bf{5.71} & 5.75 & \bf{5.55} & \bf{8.32} & 9.88 & \bf{4.59} & 13.31 & \bf{9.71} & \bf{10.49} & 11.11 & 12.69 & \bf{14.66} & 13.75 & \bf{11.72} & \bf{19.82} & \bf{20.94} \\
54 & \bf{5.04} & 6.21 & 6.30 & 6.23 & 6.09 & \bf{5.57} & 5.85 & 8.47 & \bf{9.69} & 4.60 & \bf{12.89} & 10.17 & 10.59 & \bf{10.70} & \bf{12.10} & 15.10 & \bf{13.58} & 12.57 & 25.61 & 24.22 \\
\specialrule{.1em}{.05em}{.05em}
\end{tabular}
\vspace*{-3mm}
\caption{\textbf{Per-joint rotation error comparison of the 108-vertices and 54-vertices models.} 3D heatmaps with $16\times{}16\times{}16$ resolution are used in both experiments in this table.}
\vspace*{-3mm}
\label{table:angular_dist_108vertices_54vertices}
\end{table*}
%%%%%%%%%%%%%%%%%%%%%%%%%%%%%%%%%%%%%%%%%%%%%%%%%%%%%%%%%%%%%%%%%%%%%%%%%%%%%%%%%%%%%%%%%%%%%%%%%%

The 54 vertices model shows worse MPVE and angular distance performance compared to other sub-vertices models because the number of vertices on the arms and hands is relatively small. Too few vertices do not provide enough information to resolve ambiguity in joint rotation and shape estimation. Consequently, the 54 vertices model results in higher wrist rotation errors than the 108 vertices model, which is presented in Table~\ref{table:angular_dist_108vertices_54vertices}. A visualization of the vertex positions of the 54 vertices model and other sub-vertices models is presented in Fig.~\ref{fig:sub-vertices location}.

%-------------------------------------------------------------------------
\section{Single-view Method for Visibility}

%%%%%%%%%%%%%%%%%%%%%%%%%%%%%%%%%%%%%%%%%%%%%%%%%%%%%%%%%%%%%%%%%%%%%%%%%%%%%%%%%%%%%%%%%%%%%%%%%%
\begin{table}[t]
\centering
{\small
\begin{tabular}{L{2.8cm}|C{1.2cm}|C{1.2cm}|C{1.3cm}}
\specialrule{.1em}{.05em}{.05em}
{Single-view method} & {MPJPE} $\downarrow$ & {MPVE} $\downarrow$ & {Angular} $\downarrow$ \\ 
\hline
I2L-MeshNet~\cite{2020_I2L} & \bf{17.59}  & \bf{23.70} & \bf{11.33}  \\ 
\cellcolor{Gray}METRO~\cite{2021_Lin_CVPR} & \cellcolor{Gray}18.15  & \cellcolor{Gray}23.98 & \cellcolor{Gray}11.55  \\ 
Graphormer~\cite{2021_Lin_ICCV} & 17.77  & 24.23 & 11.52  \\ 
\specialrule{.1em}{.05em}{.05em}
\end{tabular}
}
\vspace*{-3mm}
\caption{\textbf{Ablation results on the single-view mesh reconstruction method in the visibility module.}}
\label{tab:single-view method for visibility}
\vspace*{-1mm}
\end{table}
%%%%%%%%%%%%%%%%%%%%%%%%%%%%%%%%%%%%%%%%%%%%%%%%%%%%%%%%%%%%%%%%%%%%%%%%%%%%%%%%%%%%%%%%%%%%%%%%%%

In this section, we present justification for the use of I2L-MeshNet~\cite{2020_I2L} in the proposed visibility module. To this end, we construct three visibility modules by combining three state-of-the-art methods for single-view human mesh reconstruction (i.e., I2L-MeshNet, METRO~\cite{2021_Lin_CVPR}, and Graphormer~\cite{2021_Lin_ICCV}) and a visibility computation algorithm~\footnote{https://github.com/MPI-IS/mesh}. A detailed procedure for visibility estimation based on single-view mesh reconstruction is presented in Sec.~3.2 of the main paper. Table~\ref{tab:single-view method for visibility} shows the performance comparison for the cases in which three visibility modules are used in the proposed method. We found that using I2L-MeshNet produces better results than using other methods. Based on this result, we adopt I2L-MeshNet in our visibility module.

%-------------------------------------------------------------------------
\section{Multi-view Inconsistency}

\begin{figure}[t]
\centering
\includegraphics[width=\linewidth]{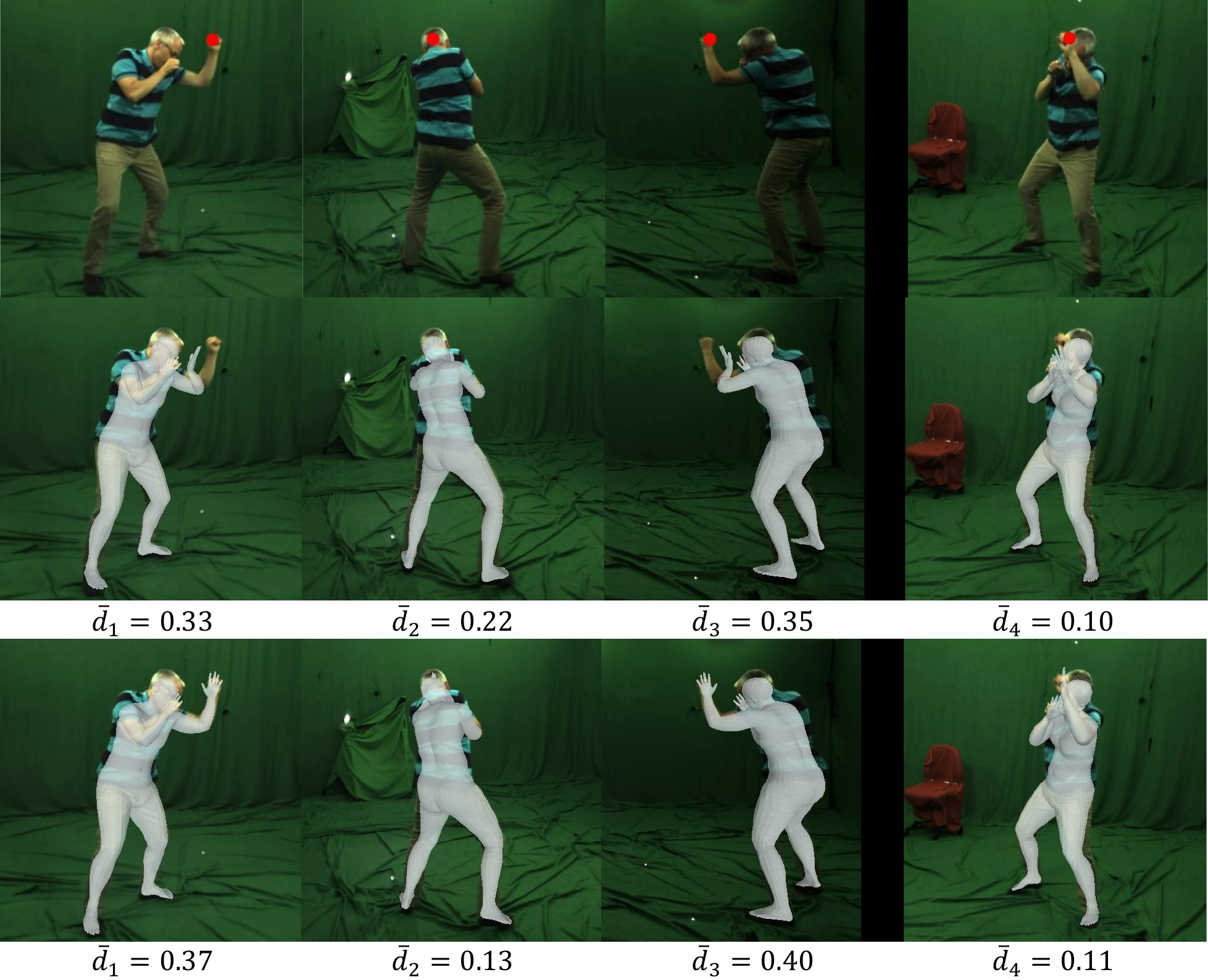}
\vspace*{-4mm}
\caption{The first row visualizes the input multi-view images. The second and third rows show the reconstructed meshes generated from the softmax baseline and LMT, respectively.}
\label{fig:multi-view inconsistency 1}
\vspace*{-4mm}
\end{figure}

\begin{figure}[t]
\centering
\includegraphics[width=\linewidth]{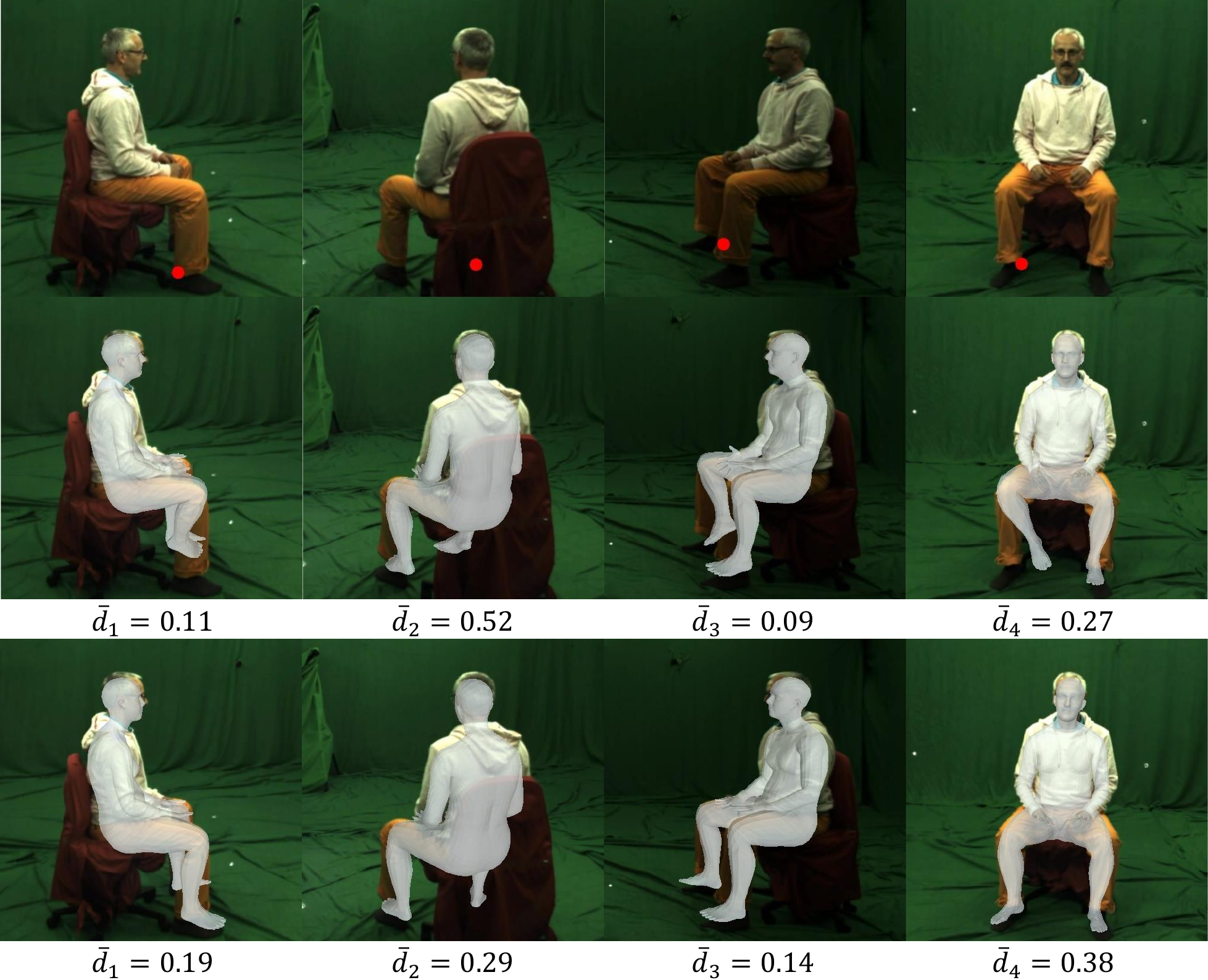}
\vspace*{-4mm}
\caption{The first row visualizes the input multi-view images. The second and third rows show the reconstructed meshes generated from the softmax baseline and LMT, respectively.}
\label{fig:multi-view inconsistency 2}
\vspace*{-4mm}
\end{figure}

This section presents additional examples on multi-view inconsistency in the ablation experiments of the main paper. Fig.~\ref{fig:multi-view inconsistency 1} gives a scenario where the left hand is invisible due to occlusion. In the second view, it is difficult to determine the exact position of the left hand because the subject's left hand is not visible. However, in the remaining views, the position of the subject's left hand can be easily found. Therefore, in order to reconstruct the left-hand mesh, the model is desirable to have a higher dependence on the features obtained from the remaining views other than the second view. However, the softmax baseline has a relatively high dependence on the features obtained from the second view. As a result, the softmax baseline incorrectly reconstructs the left hand. However, LMT reduces the dependence on the features obtained from the second view and successfully reconstructs the human mesh.

Fig.~\ref{fig:multi-view inconsistency 2} shows a scenario where the subject's right foot cannot be seen well in the second view. According to the results, the softmax baseline fails to reconstruct the mesh, but LMT reconstructs it successfully. Similar to the case of Fig.~\ref{fig:multi-view inconsistency 1}, it can be seen that the use of visibility reduces the dependence on the feature obtained from the second view where occlusion occurs.

%-------------------------------------------------------------------------
\section{Qualitative Results}

\begin{figure*}[t]
\centering
\includegraphics[width=\linewidth]{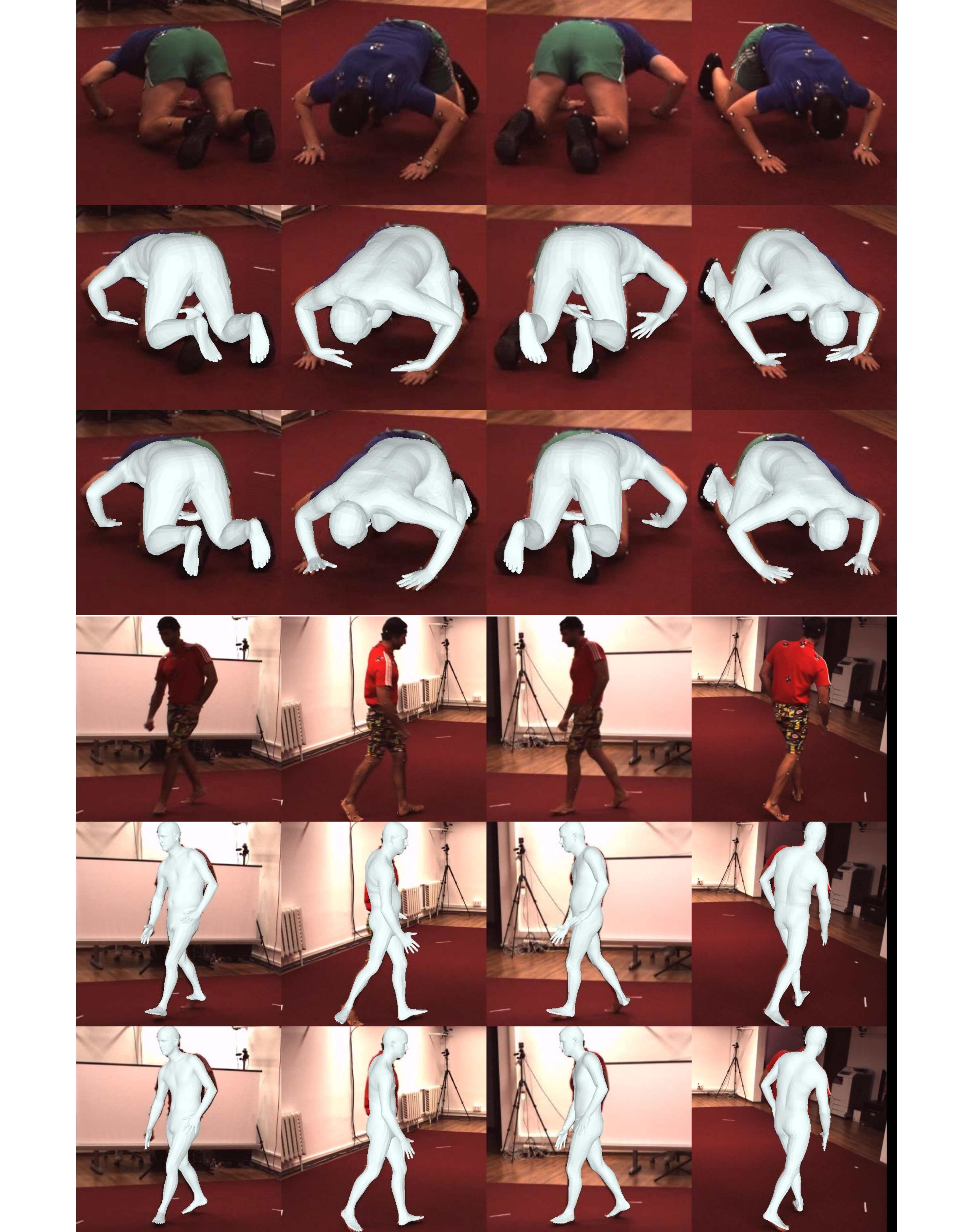}
\vspace*{-4mm}
\caption{\textbf{Qualitative results on Human3.6M.} The first and fourth rows show the input images. The second and fifth rows visualize the human meshes reconstructed by LT-fitting. And the third and sixth rows visualize the human meshes reconstructed by LMT.}
\label{fig:qualitative h36m}
\vspace*{-4mm}
\end{figure*}

\begin{figure*}[t]
\centering
\includegraphics[width=\linewidth]{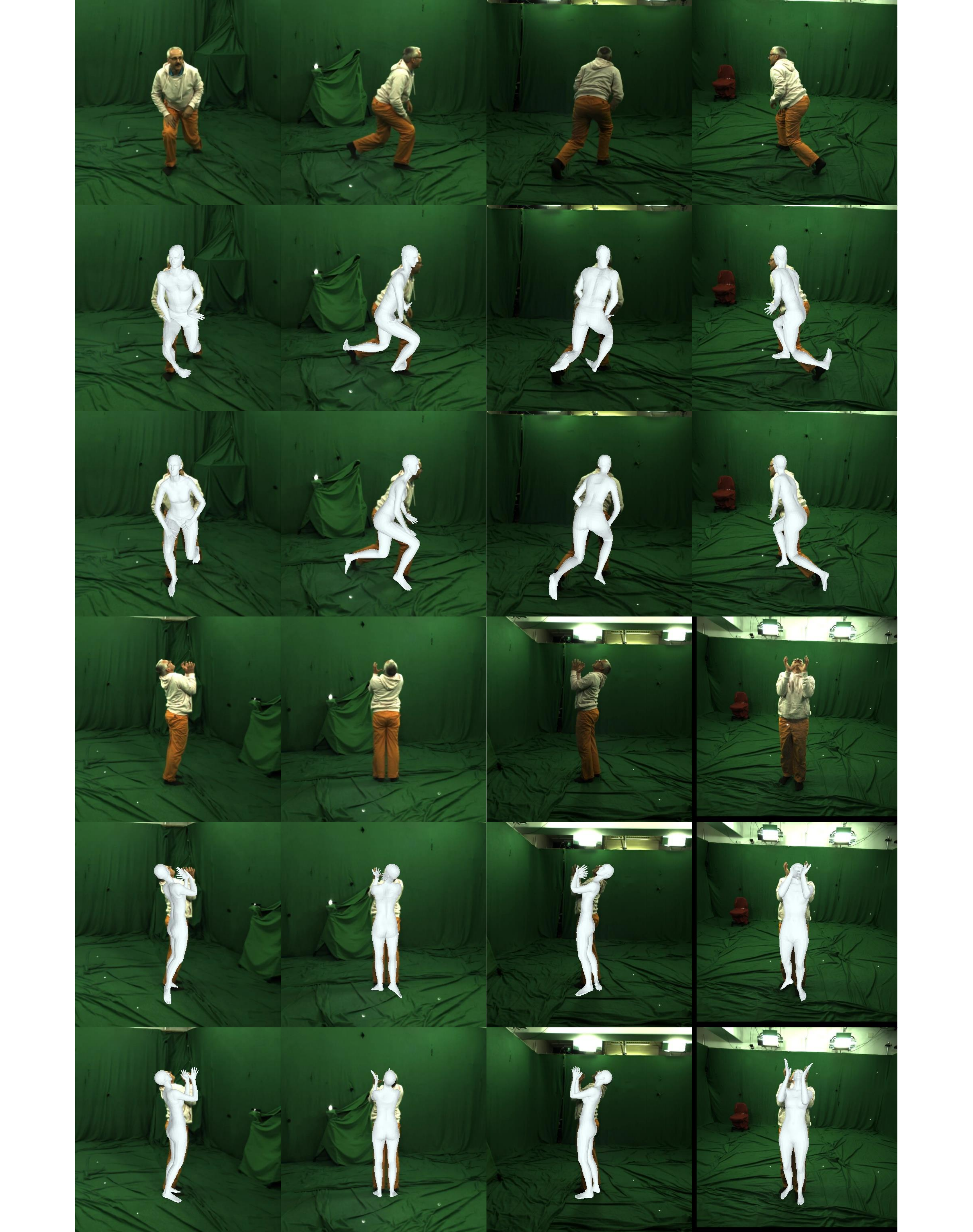}
\vspace*{-4mm}
\caption{\textbf{Qualitative results on MPI-INF-3DHP.} The first and fourth rows show the input images. The second and fifth rows visualize the human meshes reconstructed by LT-fitting. And the third and sixth rows visualize the human meshes reconstructed by LMT.}
\label{fig:qualitative mpiinf3dhp}
\vspace*{-4mm}
\end{figure*}

This section shows that our surface fitting produces qualitatively better results in terms of joint rotation and shape compared to joint fitting~\cite{2019_LT, lightcap2021}. Fig.~\ref{fig:qualitative h36m} shows the human meshes reconstructed by LT-fitting and LMT on the Human3.6M~\cite{2014_H36M} dataset. The second row of Fig.~\ref{fig:qualitative h36m} shows that LT-fitting incorrectly predicts the rotations of the left ankle, elbows, and wrists. The fifth row of Fig.~\ref{fig:qualitative h36m} shows that LT-fitting incorrectly reconstructs the right knee rotation and human shape. However, in both cases, LMT accurately predicts joint rotation and human shape.
% Fig.~\ref{fig:qualitative h36m} shows that LT-fitting incorrectly predicts wrist and elbow rotations.
% The second row of Fig.~\ref{fig:qualitative mpiinf3dhp} shows that LT-fitting incorrectly predicts the rotations of the right shoulder, elbows, wrists, knees, and ankles.

Fig.~\ref{fig:qualitative mpiinf3dhp} shows the human meshes reconstructed by LT-fitting and LMT on the MPI-INF-3DHP~\cite{2017_3DHP} dataset. The second row of Fig.~\ref{fig:qualitative mpiinf3dhp} shows that LT-fitting incorrectly predicts the rotations of the right shoulder, elbows, wrists, knees, and ankles. The fifth row of Fig.~\ref{fig:qualitative mpiinf3dhp} shows that LT-fitting incorrectly predicts the rotations of the neck, wrists, elbows, and right knee. However, similar to the results on Human3.6M, LMT accurately predicts joint rotations in both cases. As can be seen from Figs.~\ref{fig:qualitative h36m} and \ref{fig:qualitative mpiinf3dhp}, the human mesh reconstructed with accurate joint rotation and shape information can explain the human body more naturally, and we qualitatively prove the superiority of surface fitting based on these results.

{\small
\bibliographystyle{ieee_fullname}
\bibliography{LMT}
}

\end{document}